\documentclass{article}

\usepackage{algorithm2e}
\usepackage{comment}

\usepackage[preprint]{neurips_2022}

\usepackage[utf8]{inputenc} % allow utf-8 input
\usepackage[T1]{fontenc}    % use 8-bit T1 fonts
\usepackage{hyperref}       % hyperlinks
\usepackage{url}            % simple URL typesetting
\usepackage{booktabs}       % professional-quality tables
\usepackage{amsfonts}   % blackboard math symbols
\usepackage{amsmath}
\usepackage{amssymb}
\usepackage{subfigure}
\usepackage{mathtools}
\usepackage{amsthm}
\usepackage{nicefrac}       % compact symbols for 1/2, etc.
\usepackage{microtype}
\usepackage[pdftex]{graphicx}% microtypography
\usepackage{xcolor} % colors

\usepackage{bm}

\newcommand{\vM}{{\bm{M}}}

\newcommand{\vW}{{\bm{W}}}
\newcommand{\vQ}{{\bm{Q}}}

\newcommand{\vF}{{\bm{F}}}

\newcommand{\saf}{{\textsc{saf}}}

\title{Coordinating Policies Among Multiple Agents\\via an Intelligent Communication Channel}

\author{Dianbo Liu \\
  Mila-Quebec AI institute\\
  \texttt{Dianbo.liu@mila.quebec} \\
  \And
  Vedant Shah \\
  Mila and Birla Institute of Technology and Science\\
  \texttt{vedant.shah@mila.quebec} \\
  \And
  Oussama Boussif \\
  Mila-Quebec AI institute\\
  \texttt{oussama.boussif@mila.quebec} \\
  \And
  Cristian Meo \\
  TUDelft\\
  \texttt{C.Meo@tudelft.nl} \\
    \And
  Anirudh Goyal \\
  Mila-Quebec AI institute\\
  \texttt{anirudhgoyal9119@gmail.com} \\
    \And
  Tianmin Shu \\
  MIT\\
  \texttt{tshu@mit.edu} \\
    \And
  Michael Mozer \\
  Google\\
  \texttt{mcmozer@google.com} \\
    \And
 Nicolas Heess  \\
  DeepMind\\
  \texttt{heess@google.com} \\
      \And
 Yoshua Bengio  \\
   Mila-Quebec AI institute\\
  \texttt{yoshua.bengio@mila.quebec} \\

}

\begin{document}

\maketitle

\begin{abstract}
%Multi-Agent Reinforcement Learning (MARL) has seen a surge in methods that incorporate explicit mechanisms to exchange information between different agents either by pairwise communication wherein every pair of agents exchange relevant information or by using targeted communication wherein different agents learn both what to communicate and whom to communicate with. In this paper, we propose an alternative approach whereby agents communicate with each other using a stateful and active communication faciliator (\emph{\saf}) that adapts and learns to improve the collective performance of the agents. To ensure independence among agents, they are incentivized to minimize the influence of \emph{\saf} on their behaviour.  To improve adaptivity, we introduce a pool of policies  that capture the patterns shared across different agents in different contexts. At each time-step, each agent can choose a policy based on its current state and information made available from \emph{\saf}.  We demonstrate our approach's strength over existing baselines on several cooperative MARL environments. 

In Multi-Agent Reinforcement Learning (MARL), specialized channels are often introduced that allow agents to communicate
directly with one another. In this paper, we propose an alternative approach whereby agents communicate through an intelligent facilitator that learns to sift through and interpret signals provided by all agents to improve the agents' collective performance. To ensure that this facilitator does not become a centralized controller, agents are incentivized to reduce their dependence on the messages it conveys, and the messages can only influence the selection of a policy from a fixed set, not instantaneous actions given the policy. We demonstrate the strength of this architecture over existing baselines on several cooperative MARL environments. 
\end{abstract}

%To ensure that the different agents don't overfit to the information from the knowledge source and to improve the generalization performance, we train agents that in addition to maximizing reward, minimize the dependence of the selected policy on the information from the shared knowledge source. 

%We incentivize agents to learn task structure by training policies that perform well under a variety of goals, while not overfitting to any individual goal.
%We propose a targeted communication architecture for multi-agent reinforcement learning, where agents learn both what messages to send and whom to address them to while performing cooperative tasks in partially-observable environments.

%\section{Story.}

%\begin{itemize}
%    \item Multiple agents interacting.
%    \item Coordination via a shared knowledge source.
%    \item Flexibility to read information via a mixture policy.
%    \item Minimize dependence of actions on the shared knowledge source.
%\end{itemize}

%(1) is shared knowledge (SK) and/or agent's state (AS) used to select policy.
%(2) is SK and/or AS as input to the selected policy
%(3) is SK and/or AS used to update agent's state
%(4) when both shared knowledge and agent's state are affecting some aspect of processing, are those two vectors added or concatenated?  
%(5) is the KL term applied to the action distribution or the gating-network outputs?

\section{Introduction}
%Many applications in the real world---as well as in other complex, dynamic environments---require the participation
%of more than one agent, e.g.,  robotics and video game applications.
%Many applications in the real world, like robotics, but also other complex dynamic environments, such as video games, require the participation of more than one agent.
%The participation of multiple agents is required in many real world applications of RL such as robotics, as well as in other complex dynamic environments, such as video games.
Multi-agent reinforcement learning (MARL) addresses the sequential decision-making of two or more autonomous 
agents that operate in a common environment, each of which aims to optimize its own long-term return 
by interacting with the environment and with other agents \citet{busoniu2008comprehensive}. 
MARL is becoming more common in many real-world applications such as robotics, as well as in other applications involving
complex, dynamic environments, such as video games.
Largely, 
MARL algorithms can be placed into two categories depending on the type of setting they address.
In the \emph{cooperative} setting, agents collaborate to optimize a shared long-term return
whereas in the \emph{competitive} setting, an advantage for one agent results in a loss for another.

%Recent years have witnessed significant advances in reinforcement learning (RL),
%which has registered tremendous success in solving various sequential decision-making problems in machine learning. Many of the successful RL applications, e.g. robotics and video games, involve the participation of more than one single agent, which naturally fall into the realm of multi-agent RL (MARL). MARL addresses the sequential decision-making problem of multiple autonomous agents that operate in a common environment, each of which aims to optimize its own long term return by interacting with the environment and other agents \citet{busoniu2008comprehensive}. Largely, MARL algorithms can be placed into three groups: fully cooperative, fully competitive, and a mix of the two, depending on the types of settings they address. 
%In this study, we focus on cooperative MARL with partial observation. In the cooperative setting, agents collaborate to optimize a joint long-term return. In this setting, each agent only has an incomplete view of the environment and may want to exchange information with other agents to gain additional information to drive their decision-taking and obtain rewards. 

Early research in cooperative MARL focused on agents that operated independently and that did not explicitly communicate \cite{IPPO, IAC, IQL}. However, when each agent has only a partial view of the environment, agents benefit from exchanging information with one another, allowing
them to construct a more complete belief state and improve decision making.
Even in fully observed environments, inter-agent communication can be beneficial to coordinate behavior.
Not surprisingly, performance advantages are obtained when agents have the ability to learn a communication protocol, whether implicitly 
or explicitly \cite{qmix,vdn,lowe2017multi,IPPO}. 
Recent research on communication for deep MARL adopts an end-to-end training procedure based on a differentiable communication channel 
\citet{sukhbaatar2016learning, foerster2016learning, gupta2017cooperative, singh2018learning, das2019tarmac}.  
Essentially, each agent has the capability of generating messages and these messages can influence other agents' policies.
In these works, the message-generation subnet is trained using the gradient of other agents' policy or critic losses.

In designing a MARL architecture, a key decision involves the nature of the communication channel.
The simplest scenario involves direct agent-to-agent communication, where each of the $N$ agents can receive messages from all other 
agents.  Communication costs are quadratic in $N$ and each agent faces the challenge of interpreting $N-1$ simultaneous messages.
Communication costs can be reduced by restricting the communication topology \citet{wang2017nonlocal}. 
Message processing can be simplified using a learned key-value attention mechanism that condenses messages at either the side
of the sender \citet{kim2020communication} or recipient \citet{das2019tarmac}.

\textbf{Intelligent communication channel.}
In previous approaches, the communication channel is \emph{passive}, by which we mean that its role is to convey whatever message passes through the channel without alteration.  We describe an approach in which the communication channel is 
\emph{active} in that it can interpret and transform signals from one or more agents, and it is \emph{stateful} in  that its interpretations can depend on the recent history of messages transmitted.  Because we endow the channel itself 
with intelligence, message communication complexity is reduced from quadratic to linear in $N$.
In essence, the channel is a specialist agent aiming to \emph{facilitate} coordination among the other agents.
We refer to the channel as \emph{\saf}, an acronym on \emph{stateful, active facilitator}.
\saf\ is itself adaptive and learns to improve the collective performance of the agents. 

\begin{figure}
  \centering
  \includegraphics[width=\linewidth]{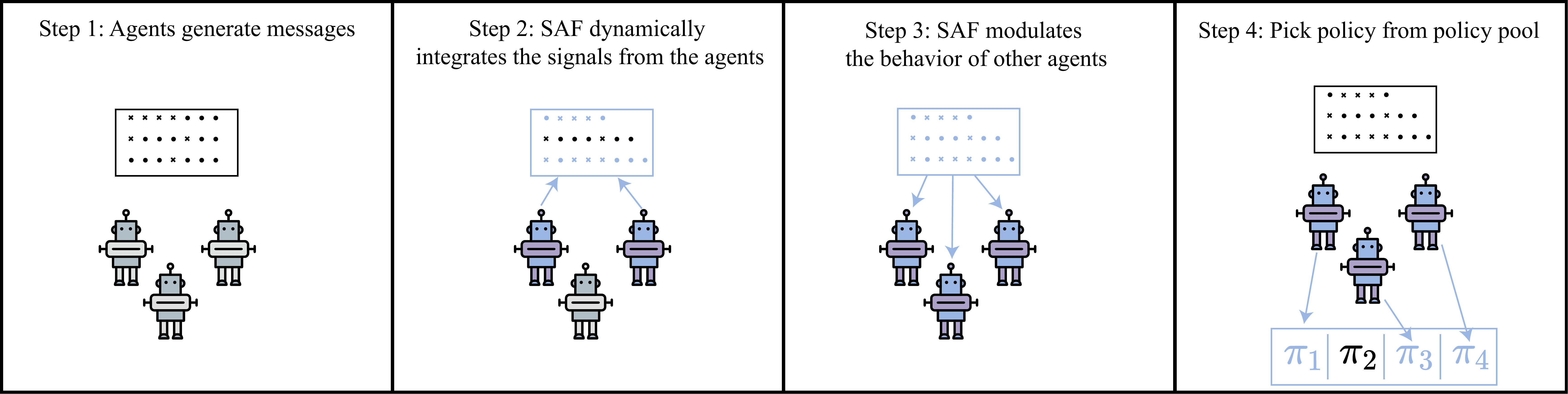}
  \caption{Agents communicate through the \emph{\saf} and pick a policy from the shared policy pool. First, each agent generates a message and competes for write-access into the \emph{\saf}. Next, all agents read messages from the \emph{\saf} and use it with their internal state to pick a policy from the shared policy pool.}
    \label{fig:illust1}
\end{figure}

\textbf{Maximizing agent independence.}
By endowing the communication channel with intelligence, there is a risk that \saf\ may simply become a centralized controller 
dictating actions to the agents, which is antagonistic to a multi-agent architecture. We therefore need to encourage 
independence of the agents, which could also lead to specialization of labor and thus faster learning. To the extent that independent agents can solve a task, independence is a clear advantage because the 
learning problem can be decomposed into smaller problems, i.e., each agent can learn without concern about the behavior of other agents.  However, most tasks require \emph{some} coordination among agents. We therefore pursue an approach 
that attempts to regularize toward independence: agents pay a penalty for modulating their behavior based on the information 
provided by \saf. This penalty, which is added to the primary task-based reward, discourages unnecessary use of the 
communication channel.

%\textbf{Minimizing the dependence of actions on shared knowledge source.} We want different agents to perform well by communicating with each other via \emph{\saf}. One may imagine that the learning may be easier if there is a solution that does not require any communication at all i.e., every agent can learn what to do without having to worry about what the other agents do. At the same time, different agents need to communicate information to solve the task at hand. Hence, we should prefer such solutions that are more modular  or more generally those that require as few bits of communication between the agents as possible (i.e, promotes independence among agents' learning tasks). We propose to achieve this by training agents that in addition to maximizing reward, minimize each agent's dependence on the information from \emph{\saf}, quantified by the conditional mutual information $I\!\left(A;M \mid S\right)$ between the agent's action $A$ and the information $M$ read from the \emph{\saf}, given the current local agent state $S$ (either the current partial observation / or the history of partial observations). This approach of learning can be interpreted as encouraging different agents to follow a \emph{default behavior} i.e., the policy which the agent should follow in the absence of any other information (i.e., from the \emph{\saf} fed by other agents). 

The penalty is expressed as the conditional mutual information between an agent's behavior, $B$, and the information, $M$,
obtained from \saf, given the agent's current belief state, $S$, denoted $I\!\left(B; M \mid S\right)$. 
Previous work has shown how to optimize $I\!\left(B;M \mid S\right)$ using the framework of KL-Regularized RL  \citet{teh2017distral,  galashov2019information, goyal2019infobot, tirumala2020behavior}. 
This optimization encourages the agent to act according to a \emph{default} or \emph{prior} policy that is insensitive to \saf.
% HOW ABOUT:
 To see that minimizing $I\!\left(B;M \mid S\right)$ is achieved by minimizing the KL divergence between an agent's policy
 and the default one, note that
%To see that minimizing $D_{\text{KL}}[policy||prior]$ minimizes the MI relationship, note that the regularizer can also be written as 
$I\!\left(B;M \mid S\right) = \mathbb{E}_{\pi_\theta}\left[D_{\text{KL}}\!\left[\pi_\theta \!\left(B \mid S,M\right) \mid \pi_0 \!\left(B \mid S\right) \right]\right]$, 
where $\pi_\theta \!\left(B \mid S,M\right)$ is the agent's \saf-sensitive policy,
$\pi_0 \!\left(B \mid S\right)= \pi_\theta \!\left(B \mid S\right)$  is the default policy,
$\mathbb{E}_{\pi_\theta}$ denotes an expectation over trajectories generated by $\pi_\theta$, and
$D_{\text{KL}}$ is the  Kullback-Leibler divergence.
In past work, the information asymmetry obtained by prohibiting the default policy access to shared information
has been shown to improve transfer and generalization
\citet{galashov2019information, goyal2019infobot, tirumala2020behavior}. 

\textbf{Policy switching}. We were deliberately vague in describing $B$ as an agent's behavior.  $B$ might refer to the actions an agent takes \citet[as it has in][]{goyal2019infobot, galashov2019information, tirumala2020behavior}, but in the present work, we endow agents with a discrete set of distinct policies to select from, and the agents have an explicit decision-making
component that selects a policy stochastically conditioned on $S$ and $M$.  The communicated information $M$ is used only
to select a policy, not to select actions conditioned on the policy. This mixture-of-policies approach 
limits the manner in which \saf\ can micromanage an agent's behavior, and it has been shown to be effective in
endowing agents with different behavior modes \citet{goyal2021scoff,tirumala2020behavior}.

In the previous paragraphs, we introduced three key ideas that operate synergistically. First, communication among agents is via an intelligent channel, \saf. Second, each agent is incentivized to act independently and avoid relying on \saf. Third, agents
operate according to a mixture-of-policies, where \saf\ provides the signal to select the policy.
To better appreciate how these ideas work in tandem, consider the naturalistic example of a herd of deer coordinating their
behavior. By default, each animal's policy is to graze in a field. But when one of the animals senses danger, the herd
needs to escape. They need to move in unison because if they split up, it will be easier for a predator to trap them.
Suppose that they can escape either to the north or the east, each characterized by a different policy. \saf\ in
this case will relay the danger alert and will collect information from the individuals and suggest a direction so that
the deer can escape in coordination. Each deer is responsible for avoiding obstacles and avoiding running into other deer,
and thus they operate autonomously with only the high-level guidance from \saf.

\vspace{-3mm}
\paragraph{Contributions.} The key contributions of our work are as follows: (a) we propose a novel architecture for multi-agent RL. Instead of agents communicating directly between one another, communication is mediated by a facilitator, \saf, which itself uses the history of communication and active computation to improve the collective performance of all the agents, (b) To ensure agent autonomy, different agents are incentivized to minimize the influence of \saf\ on their behavior, (c) To further promote autonomy, agent behavior is only coarsely modulated by \saf, much in the way that a Ph.D. student's research direction is only guided at a high level by their advisor. In the case of our MARL architecture, this modulation  comes in the form of policy selection, which is made explicit via a policy mixture model with a component that switches among policies (see Figure \ref{fig:illust1}), (d) We show the performance of the proposed method on different MARL environments in cooperative setting. We also conduct various ablations and understand the role of different components namely the use of intelligent channel \emph{\saf}, maximizing agent independence and the use of policy pool. We show that the collective performance of the agent which uses all the three components is higher as compared to an agent which only uses one of the components.

\section{Background and Notation}
In this work, we consider a multi-agent version of Markov decision processes (MDPs) with partially observable Markov environments (POMDP).
The environment is defined as $(\mathcal{N},\mathcal{S},\mathcal{O},\mathcal{A},\Pi,R,\gamma)$. $\mathcal{N} = \{1,...,N\}$ denotes the set of $N>1$ agents. $\mathcal{S}$ describes all possible states of all agents.  $\mathcal{A} = A_1\times\dots\times A_N$ denotes the joint action space and $a_{i,t}\in A_i$ refers to the action of agent $i$ at time step $t$. $\mathcal{O}=O_{1}\times\dots\times O_{N}$
denotes the set of partial observation where $o_{i,t}\in O_i$ stands for partial observation of agent $i$ at time step $t$. $\Pi$ is the set of policies available to the agents. To choose actions, agent $i$ uses a stochastic policy $\pi_{\theta_i}:O_i \times A_i \mapsto [0,1]$. Actions from all agents together produces the transition to the next state according to transition function $\mathcal{T}:(s_t, a_{1,t}, \dots, a_{N,t}) \mapsto s_{t+1}$ where $s_t\in\mathcal{S}$ is the joint state at timestep $t$. 
$R:\mathcal{S}\times\mathcal{A}\mapsto \mathbb{R}$ is the global reward function conditioned on the joint state and actions. At timestep $t$, the agent team receives a reward $r_t=R(s_t,a_{1,t},\dots,a_{N,t})$ based on the current total state $s_t$ and joint action $a_{1,t},\dots, a_{N,t}$. $\gamma$ is the discount factor for future rewards. In this study, we focus on cooperative MARL with partial
observations.

\section{MARL with a Facilitator}

We present our approach, first in terms of a high level overview of the mechanisms enabling inter-agent communication
(Section~\ref{sec:hlo}), and then with a detailed description that steps through the algorithm (Section~\ref{sec:da}).
For further details, see Algorithm \ref{alg:attention} in Appendix~\ref{appx:alg}.
%Here we first give a detailed description of our approach and then present the algorithm in detail.
%We begin with an overview of the main three mechanisms for inter-agent communication.
%in the partial observation setting. 

\subsection{High Level Overview}
\label{sec:hlo}

%%F:  the state of the slots.
%M': the message by the agents.
%M: the message *read* by the agents.

 \textbf{Stateful active facilitator.} The facilitator, \emph{\saf}, consists of 
$l$ stateful memory \emph{slots}, each a vector of $d_m$ elements. The slots are randomly initialized at the beginning of an episode. The state of the \emph{\saf} is updated once per time step, where a time step corresponds to all agents performing one action. In this update, \emph{\saf} integrates information provided by the agents into its slot memory and then
outputs a message $M$ that any agent can use.

%At each computational time-step $t$, \emph{\saf} is integrating the information from the agents and weighing the information that it finds interesting to update it's state, and then \emph{\saf}  modulates the behavior of other agents. 

\textbf{Policy switching using a shared policy pool.} Although all agents operate in the same environment, 
their contexts and objectives may vary. A policy pool $\Pi = \{ \pi^1, \ldots, \pi^U\}$, shared among agents, 
enables agents to exhibit diverse behaviors and have distinct goals. Each policy $\pi^u \in\Pi$ is associated with a 
learned signature key  $k_{\pi^u}$. Using differentiable hard attention with  Gumbel-softmax \citet{jang2016categorical},
an agent can  dynamically select one of  the policies at each time step via a query formed from its belief state, $S$, 
and the message conveyed by \saf, $M$. 

\textbf{Reducing agent dependence on \saf.} To reduce an agent's dependence on \saf, each agent is penalized
according to the KL divergence $D_\text{KL} [ \Pr(Z \mid S,M) \mid\mid \Pr(Z \mid S) ]$, where $Z$ is the agent's policy choice.

%\begin{aligned}
%\label{objective}
%J\!\left(\theta\right) &\equiv \mathbb{E}_{\pi_\theta}\!\left[r\right]-\beta I\!\left(Z; M\mid S\right) \\
%&= \mathbb{E}_{\pi_\theta }\!\left[r - \beta D_{\text{KL}}\left[\pi_{\theta}\!\left(Z \mid S, M\right) \mid \pi_0\!\left(Z \mid S\right)\right]\right],
%\end{aligned}
%\end{equation}
%where $\mathbb{E}_{\pi_\theta}$ denotes an expectation over trajectories across different agents, $\beta > 0$ is a trade-off parameter and $r=\sum_{t=0}^T r_t$ is the total return up to the horizon $T$.
%\fi
%, $D_{\text{KL}}$ is the Kullback–Leibler divergence, and $\pi_0 \!\left(Z \mid S\right)\equiv \sum_g p\!\left(g\right) \pi_\theta \!\left(Z \mid S,g\right)$ is a ``default'' policy with the agent's goal marginalized out.

%Different policies in the shared pool can be seen as "different behaviors"  to deal with different contexts and scenarios. 

%Given the state $S$ of the agent and the information $M$ read from the shared knowledge source, each agent can independently and dynamically select a policy $\pi^k_{\theta}\!\left(A \mid S, M\right)$ from a pool of $K$ policies where $\pi^k_{\theta}\!\left(A \mid S, M\right) = \sum_z p_\text{enc}\!\left(z_{k}\mid S,M\right)  p_\text{dec}\!\left(A\mid S,z_{k}\right)$\footnote{For our experiments, we estimate the marginalization over $Z$ using a single sample.}. 
%\end{enumerate}

\subsection{Detailed Algorithm}
\label{sec:da}
\textbf{Step 1: Agents pass messages to \saf.} At step $t$, each agent $i$ receives a partial observation $o_{i, t}$ 
of the environment. This observation is used to update the agent's belief state, $s_{i, t}$, which in turn is
used to generate a message for \saf.  The message,  $m'_{i, t} = g_{enc}(s_{i, t})$ is a vector of dimensionality
${d_m}$. We denote the set of  messages generated by the agents at time step $t$ as 
$\vM'_t \equiv \{m'_{i, t} | 1 \leq i \leq N\}  \in \mathbb{R}^{N\times d_m}$.

% In order to implement the competition between specialists to write into the workspace, we  use a key-query-value attention mechanism. In this case, the query is a function of the state of the current workspace memory content, represented by matrix $\vM$ (with one row per slot of the memory), i.e $\widetilde{\vQ} = \vM \widetilde{\vW}^{q}$.  Keys and values are a function of the information from the specialists  i.e., a function of $\vR$. We apply dot product attention to get the updated memory matrix:  $\vM \leftarrow \mathrm{softmax}\left(\frac{\widetilde{\vQ}(\vR \widetilde{\vW}^{e})^\mathrm{T}}{\sqrt{d_e}}\right)\vR \widetilde{\vW}^{v}$. The use of a regular softmax to write into $\vM$ leads to a standard soft competition among different specialists to write in the shared workspace. 

%One can also use a top-$k$ softmax~\citet{ke2018sparse} to select a fixed number of specialists allowed to write in the shared workspace:  based on the pre-softmax values, a fixed number of $k$ specialists  which have the highest values are selected, and get access to write in the shared workspace. Selection with a top-$k$ softmax is a hybrid between hard  and soft selection. We denote the set of thus selected specialists as $\mathcal{F}_t$. 

\textbf{Step 2: \emph{\saf} integrates information from the agents.} 
\emph{\saf} dynamically integrates the information from all the agents and incorporates the information 
that it finds interesting into its state representation. The \emph{\saf} slot memory at time step $t$ is a set of vectors 
$F_t \in \mathbb{R}^{l \times d_m}$, each row representing one of the $l$ slots. The state of 
\emph{\saf} is updated based on agent messages, ensuring that only the important 
information is incorporated.  \saf\  achieves this via the use of query-key-value attention mechanism 
\citet{bahdanau2014neural, vaswani2017attention}.  In this case, the query is a function of \saf's state 
(a set of slots), represented by matrix $\vF_{t}$ (with one row per slot), i.e., $\widetilde{\vQ} = \vF_{t} \widetilde{\vW}^{q}$.  Keys and values are a function of the messages from individual agents. Dot product attention is applied
to obtain the updated state of the slots:  $\vF_{t} \leftarrow \mathrm{softmax}\left(\frac{\widetilde{\vQ}(\vM'_{t} \widetilde{\vW}^{e})^\mathrm{T}}{\sqrt{d_e}}\right)\vM'_{t} \widetilde{\vW}^{v}$. The use of softmax to write into $l$ slots leads to a standard soft competition among agents to influence the state of \emph{\saf}. Next, self-attention
is applied over the slots of \saf\ to obtain its updated state.

\textbf{Step 3: \emph{\saf}  modulates the behavior of other agents.} \emph{\saf} makes the updated state available to the agents should they deem to use it. We again utilize an attention mechanism to perform this reading operation. All the agents create queries $Q^s_{t} = \{q^s_{i, t} | 1 \leq i \leq N\}\in\mathbb{R}^{N\times d_e}$ where $q^{s}_{i,t}$ is query generated using the encoded partial observations of agent $i$: $q^{s}_{i, t} = W^{q}_{write}s_{i, t}$. Generated queries are matched with the keys $\kappa = F_t \vW^{e}\in\mathbb{R}^{l\times d_e}$ from the updated state of \emph{\saf} (a set of slots), forming the attention matrix
\mbox{$\alpha= \mathrm{softmax} \left( \frac{Q^s_t \kappa^T
   }{\sqrt{d_e}}\right)$}. 
The slot values generated by each slot of \emph{\saf}'s state  and the attention weights are then made available to all the agents: $$
\vM_t = \mathrm{softmax} \left( \frac{Q^s_t \kappa^T
   }{\sqrt{d_e}}\right) \vF_t W^v 
$$

Here $\vM_t=  \{m_{i, t} | 1 \leq i \leq N\}$, where $m_{i, t}$ is the message made available to the agent $i$.

%In the next step each agent reads relevant information from the shared knowledge source $M_t$. This is done via a soft-attention mechanism: $M'_t = softmax(\frac{Q^{s}_{t}M_t}{\sqrt{d_slot}})M_t$ where $M'_t$ is the set of messages $\{m'_{i, t} | 0 \leq i < N\}$ to be received by the agents and $Q^{s}_{t} = \{q^{s}_{i, t} | 0 \leq i < N\}$ where $q^{s}_{i, t}$ is query generated using the encoded partial observations of agent $i$: $q^{s}_{i, t} = W^{q}_{write}s_{i, t}$.

\textbf{Step 4: Policy switching.} The encoded partial observation of an agent $s_{i, t}$ and the information made available to  each agent $m_{i, t}$ is used to select a policy $\pi^{u}$ for that agent from the pool of policies $\Pi$ via a straight-through Gumbel-softmax attention mechanism to make a differentiable approximately one-hot selection. Each policy has an associated signature key which is initialized randomly at the start of the training: $K^{\Pi}_t=\{K^{\pi^{u}}_{t} | 1 \leq u \leq U\}$. These keys are matched against the queries computed as deterministic function of encoded partial observation and the information made available to each agent.  
$q^{policy}_{i, t} = g_{psel}(s_{i, t},  m_{i, t})$  where $ g_{psel} $ is parameterized as a neural network.

\[ \mathrm{index}_i= \mathrm{GumbelSoftmax}\left(\frac{q^{policy}_{i, t} (K^{\Pi}_t)^T}{\sqrt{d_m}}\right)\]

As a result of this attention procedure, agent $i$ selects a policy $\pi^{index_{i}}$. This operation is performed independently for each agent, i.e. each agent selects a policy from the policy pool.

\textbf{Step 5: Maximizing agent independence.}  We minimize the dependence of agent on the information made available to each agent. We do this by optimizing the conditional mutual information by upper bounding the KL and penalizing $I\!\left(Z; M\mid S\right)$ \footnote{$I\!\left(Z; M\mid S\right) \geq I\!\left(A;M\mid S\right)$ \citet{galashov2019information, goyal2019infobot} such that $\pi_{\theta}\!\left(A \mid S, M\right) = \sum_z p_\text{enc}\!\left(z\mid S, M\right) p_\text{dec}\!\left(A\mid S,z\right)$}.

Thus, we can instead maximize this lower bound on $J\!\left(\theta\right)$:
\begin{equation}
\begin{aligned}
J\!\left(\theta\right) &\geq \mathbb{E}_{\pi_\theta}\!\left[r\right]-\beta I\!\left(Z; M\mid S\right) 
%&= \mathbb{E}_{\pi_\theta }\!\left[r - \beta D_{\text{KL}}\left[p_\text{enc}\!\left(Z \mid S, M\right) \mid p\!\left(Z\mid S\right) \right]\right]
\label{eq:information_bottleneck}
\end{aligned}
\end{equation}

where $\mathbb{E}_{\pi_\theta}$ denotes an expectation over trajectories across different agents, $\beta > 0$ is a trade-off parameter and $r=\sum_{t=0}^T r_t$ is the total return up to the horizon $T$.

%We use an based loss to penalize the dependence of policy selection on the message received. This loss is calculated as the negative of KL divergence between the attention weights calculated by using $q^{policy}_{i,t}$ as the query which are denoted by $A_{i,t}$ and the attention weights calculated by using $q^{policy-S}_{i,t}$ as the query where $q^{policy-S}_{i,t} = q^{s}_{i,t}$ denoted by $A^{-S}_{i,t}$. This loss is then added to the policy loss. $KL Loss = -D_{KL}(A_{i,t}||A^{-S}_{i,t})$

\section{Related Work}

\textbf{Centralized training decentralized execution (CTDE).} These approaches are among the most commonly adopted variations for MARL in cooperative tasks. They usually involve a centralized critic which takes in global information, i.e. information from multiple agents and decentralized policies which are guided by the critic. \citet{Foerster2018CounterfactualMP} uses the standard centralized critic decentralized actors framework with a \textit{counterfactual baseline}. \citet{lowe2017multi} and \citet{Yu2021TheSE} propose the extension of single-agent DDPG and single-agent PPO respectively, to a multi-agent framework with a centralized critic and decentralized policies during training and completely decentralized execution. \citet{li2021celebrating} uses an information theory based objective to promote novel behaviours in CTDE based approaches. Value Decomposition (\cite{Sunehag2018ValueDecompositionNF}, \citet{Rashid2018QMIXMV}, \citet{wang2020qplex}, \citet{Mahajan2019MAVENMV}) approaches learn a factorized action-value function. \citet{Sunehag2018ValueDecompositionNF} proposes Value Decomposition Networks (VDN) which simply add the action-value function of each agent to get the final action value function. \citet{Rashid2018QMIXMV} uses a mixing network to combine the action-value functions of each agent in a non-linear fashion.

\textbf{Communication in MARL.} Several approaches use emergent communication protocols for MARL. Communication involves deciding the message to be shared and determining how the message sending process is implemented. \citet{foerster2016learning} and \citet{Sukhbaatar2015} have done work on learnable inter-agent communication protocols. \citet{Jiang2018LearningAC} first proposed using attention for communication where attention is used for integrating the received information as well as determining when communication is needed.  \citet{das2019tarmac} uses multiple rounds of direct pairwise inter-agent communication in addition to the centralized critic where the messages sent by each agent are formed by encoding it's partial observation, location information, etc., and the messages received by each agent are integrated into it's current state by using a soft-attention mechanism.  \citet{kim2020communication} uses intentions represented as encoded imagined trajectories as messages where the encoding is done via a soft-attention mechanism with the messages received by the agent. \citet{wang2021tom2c} trains a model for each agent to infer the intentions of other agents in a supervised manner, where the communicated message denotes the intentions of each agent. The above mentioned approaches require a computational complexity which is quadratic in the number agents. Our approach has a computational complexity which is linear in the number of agents. Moreover, we show that our approach is able to outperform several standard baselines using messages which can be computed as simply the encoded partial observations of each agent. 

\textbf{Option-critic in Multi-agent Hierarchical Reinforcement learning}
 A classical approach within the Hierarchical Reinforcement Learning (HRL) \citet{Pateria2021Hierarchical} literature is modeling agents' intentions as options \citet{Sutton1999Between}, temporally extended actions that aim to achieve subgoals in a finite time horizon. Recently, \citet{Bacon2017The} proposed an end-to-end option-critic architecture capable of learning both options and the related policy. However, despite the advantages brought by using options, due to the temporally-extended nature of options, agents' responses can be inconsistent when the environment or other agents' behaviour change. To tackle this problem, \citet{Han2019Multi} proposed a dynamical termination scheme which allows an agent to flexibly terminate its current option.
 Although both option-critic and our approach use a pool of actors, while in the former case actors model options, in the latter one actors model policies, preventing agents' inconsistent behaviours. Moreover, although within the option-critic framework the optimality of the learned hierarchical policy can be theoretically guaranteed \citet{Chakravorty2020Option}, the learned options cannot be easily transferred to other tasks \citet{Pateria2021Hierarchical}. Furthermore, many works proposed within the option-critic framework (\citet{Klissarov2017Learning},\citet{Riemer2018LearningAO},\citet{Khetarpal2020OptionsOI}) perform poorly on sparse reward tasks (\citet{Bagaria2020Option},\citet{Nachum2018Data}), while experimental results show that our approach presents comparable performances with SOTA baselines even on that case (e.g., Waterworld Environment \citet{gupta2017cooperative})

\textbf{Independent learning.} Independent Learning in MARL consists of each agent optimizing its policy locally using its observation and in the absence of any communication or centralized controller (as in CTDE). These approaches mainly consist of the extension of single-agent RL approaches to multi-agent settings where each agent learns independently using local observations, considering other agents as part of the environment. \citet{Tan1993MultiAgentRL} proposed Independent Q-Learning (IQL) where each agent independently learns it's own action-value function. \citet{Witt2020IsIL} demonstrates that PPO, when used for independent learning in multi-agent settings (called Independent PPO or IPPO) is in fact capable of beating several state of the art approaches in MARL on competitive benchmarks such as StarCraft and can hence serve as a strong baseline. %\citet{Ndousse2021EmergentSL} considers a setting with no communication and uses a model based auxiliary loss in order to enable each agent to model the policies of other agents which, otherwise is not possible without access to other agent's actions or observations (i.e. communication).

\textbf{Communication bottleneck.} With the emergence of modular deep learning architectures \citet{vaswani2017attention, goyal2021neural, scarselli2008graph, bronstein2017geometric, kipf2018neural, battaglia2018relational}, which require communication among different model components, there have been development of methods which introduce a bottleneck in this communication to a fixed bandwidth which helps communicating only the relevant information. \citet{DianboLiu2021DiscreteValuedNC} uses a VQ-VAE \citet{oord2017neural} to discretize the information being communicated. Inspired by the theories in cognitive neuroscience \citet{Baars1988-BAAACT, shanahan2006cognitive, Dehaene-et-al-2017}, \citet{goyal2021coordination} proposes the use of a generic \textit{shared workspace} which acts as a bottleneck for communication among different components of multi-component architectures and promotes the emergence of \textit{specialist} components. We use a \emph{\saf} similar to the shared workspace in which different agents compete to write information to and read information from.

\section{Experiments}
We investigate how well \emph{\saf}, policy switching, and maximizing independence work together to improve cooperative MARL. Next, we compare our method with a state-of-the-art cognitive science-inspired approach used in multi-agent communication and cooperation\citet{wang2021tom2c}. Lastly, to understand if our method can be adapted to MARL methods that send different messages among agents, we integrate our machinery into a MARL algorithm that iteratively sends hypothetical actions among cooperative agents.

\subsection{Environments}
In this section, we describe the various MARL environments which we considered for our experiments..

\begin{figure*}[htp]
    \centering
    \subfigure[GhostRun Environment]{
    \includegraphics[width=0.2\linewidth]{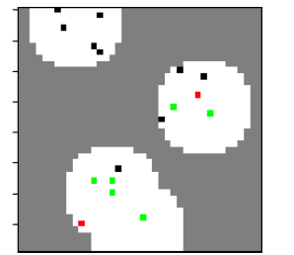}}\hspace{15mm}
    \subfigure[PistonBall Environment]{
    \includegraphics[width=0.1\linewidth]{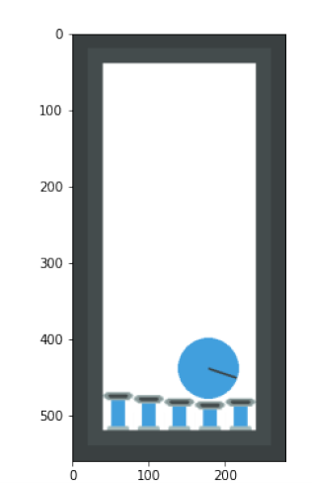}}\hspace{15mm}
    \subfigure[MSTC Environment]{
    \includegraphics[width=0.2\linewidth]{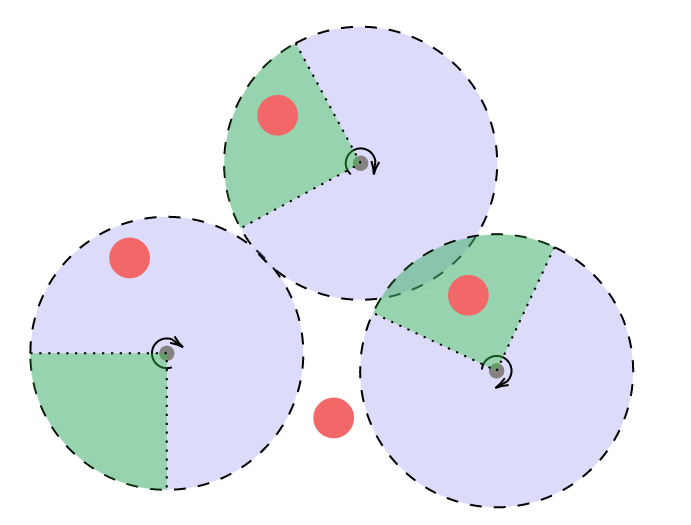}}
    
    \caption{GhostRun, PistonBall and MSTC environments. In the GhostRun environment (left panel) Each agent has its own partial view of the environment and the reward is to escape from ghosts (red dots). Different agents can communicate their encoded views with each other. In the PistonBall environment (middle panel) all agents (pistons) work together to move the ball from one side to the other. In the MSTC environment (right panel), sensors (gray dots) need to cover as many targets as possible (red dots). MSTC figure adapted from \citet{wang2021tom2c}}
    \label{fig:Env}
\end{figure*}

\vspace{-3mm}
\paragraph{GhostRun environment.} We use the \textbf{GhostRun Environment} which is an adaptation of the Drone environment available from \citet{shuo2019maenvs}. The environment consists of multiple drones, each with a partial view of the ground below them. The ground consists of ghosts - represented by red dots, trees - depicted by green dots, and obstacles - depicted by black dots. The ghosts move about randomly whereas the trees and obstacles are stationary (see Figure \ref{fig:Env} (a)). The task at hand is a cooperative task where the agents must work together to escape from ghosts. Hence, the number of ghosts in each agent's partial observation of the environment must be minimized. The reward received by each agent at each time step is the negative of the total number of ghosts in the view of all the agents and a step cost of -1 for each step taken.

\vspace{-3mm}
\paragraph{Multi-Sensor Target Coverage.} We use the multi-sensor multi-target tracking (MSTC) environment from \citet{wang2021tom2c}. There are sensors (which are the agents) and targets. The goal is for the sensors to observe as many targets as possible at once (see Figure \ref{fig:Env} (c)). Each sensor has a partial observation of its surroundings and its view may be obstructed by obstacles. The targets can move according to one of two rules: according to a random walk, or along the shortest path to reach a previously sampled destination. At the beginning of each episode, the location of targets, sensors and obstacles is randomly sampled. The maximum episode length is $100$ steps and the reward is defined as $r=\frac{1}{m}\sum_q T_q$ where $T_q=1$ if the target $T_q$ is observed by any sensor and $0$ otherwise. If no target is observed (i.e. $T_q=0\quad\forall q$) then $r=-0.1$.

\vspace{-3mm}
\paragraph{PistonBall Environment.} This is a simple physics-based cooperative game where the goal is to move the ball to the left wall of the game border by activating the vertically moving pistons. Each agent’s observation is an RGB image of the two pistons (or the wall) next to the agent and the space above them (see Figure \ref{fig:Env} (b)). Every piston can be acted on at any time. The action space can be discrete: 0 to move down, 1 to stay still, and 2 to move up. Alternatively, the action space can be continuous: the action value in the range $[-1, 1]$ is proportional to the amount by which the pistons are raised or lowered. Continuous actions are scaled by a factor of $4$, so that in both the discrete and continuous action space, the action $1$ will move a piston $4$ pixels up, and $-1$ will move pistons 4 pixels down.

%Accordingly, pistons must learn highly coordinated emergent behavior to achieve an optimal policy for the environment. Each agent gets a reward that is a combination of how much the ball moved left overall and how much the ball moved left if it was close to the piston (i.e. movement the piston contributed to). A piston is considered close to the ball if it is directly below any part of the ball. Balancing the ratio between these local and global rewards appears to be critical to learning this environment, and as such is an environment parameter. The local reward applied is 0.5 times the change in the ball’s x-position. Additionally, the global reward is change in x-position divided by the starting position, times 100, plus the time_penalty (default -0.1). For each piston, the reward is local_ratio * local_reward + (1-local_ratio) * global_reward. The local reward is applied to pistons surrounding the ball while the global reward is provided to all pistons.

\subsection{Baselines: Varying Along Different Axes}

In our experiments, we consider multiple baselines which use some aspects of the proposed method, such as the use of stateful and active facilitator \emph{\saf}, shared pool of policies and maximizing agent independence. In particular, we try to disentangle the contributions of these different components of the proposed architecture. %namely stateful and active communication channel [\emph{\saf}], shared pool of policies [\textsc{SP}]. 
The following baselines are evaluated:

\textit{Multiple Agents with no communication} [\textsc{I}]: There are multiple independent agents with no communication between them.

\textit{Multiple Agents with pairwise communication} [\textsc{P}]: 
Every pair of agents can communicate with each other via self-attention \citet{vaswani2017attention}.

\textit{Multiple Agents with pairwise communication and shared pool of policies} [\textsc{P + SP}]: Every pair of agents can communicate with each other via self-attention and each agent can select a different policy from the shared pool of policies. The difference from the proposed method would be that this variant does not make use of \emph{\saf}. 

\textit{Multiple Agents with \textsc{\emph{\saf}}} [\textsc{\emph{\saf}}]: Here, we consider multiple agents such that different agents communicate with each other via \emph{\saf}, and all agents share the same policy.

\textit{Multiple Agents with \textsc{\emph{\saf}} and policy pool} [\textsc{\emph{\saf} + SP}]: the  agents can communicate with each other via \emph{\saf}, and also share the pool of policies such that each agent can dynamically select a policy from the policy pool. 

\textit{Multiple Agents with \textsc{\emph{\saf}}, policy pool and maximizing agent dependence.} [\textsc{\emph{\saf} + SP + KL}]: The agents can communicate with each other via \emph{\saf}, share the pool of policies, and also agents minimize the dependence of the policy on the information made available by \emph{\saf}.

%such that each agent can dynamically select a policy from the policy pool. 

%In this section we compare our approach against other MARL algorithms in the GhostRun, the Multi-Sensor Target Coverage (MSTC) and the PistonBall environments. 

\vspace{-4mm}
\paragraph{Hyperparameters.}  We consider two hyper-parameters in the proposed model: (1) number of policies in the pool ($U=N_{policies}$), (2) number of slots in the \emph{\saf} ($l=N_{slots}$).  For all our experiments, we use $N_{policies}=3$.

%We experiment with $2 \leq N_{policies}, N_{slots} \leq 6$ and test it on the GhostRun environment. We find that the performance to be particularly robust against variation in the number of slots $N_{slots}$, and number of policies.

\begin{figure*}[ht!]
    \centering
    \subfigure[2 agents]{
    \includegraphics[width=0.3\linewidth]{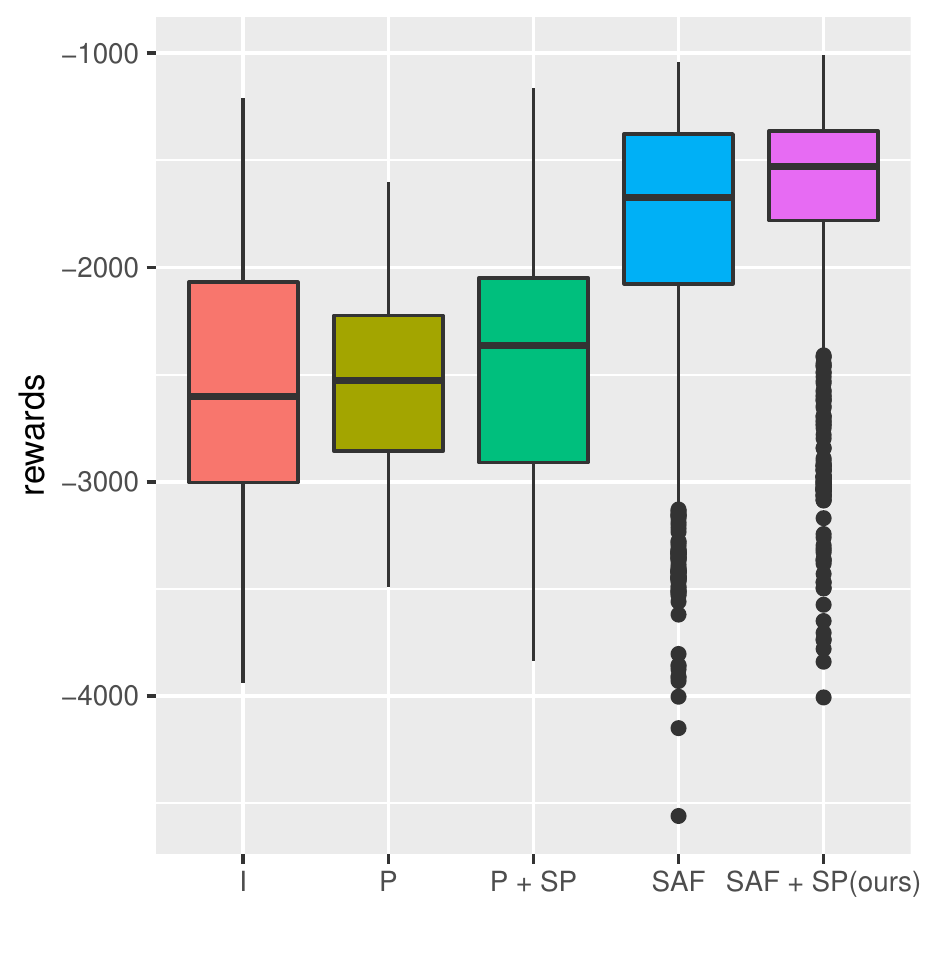}
    }
    \subfigure[5 agents]{
    \includegraphics[width=0.3\linewidth]{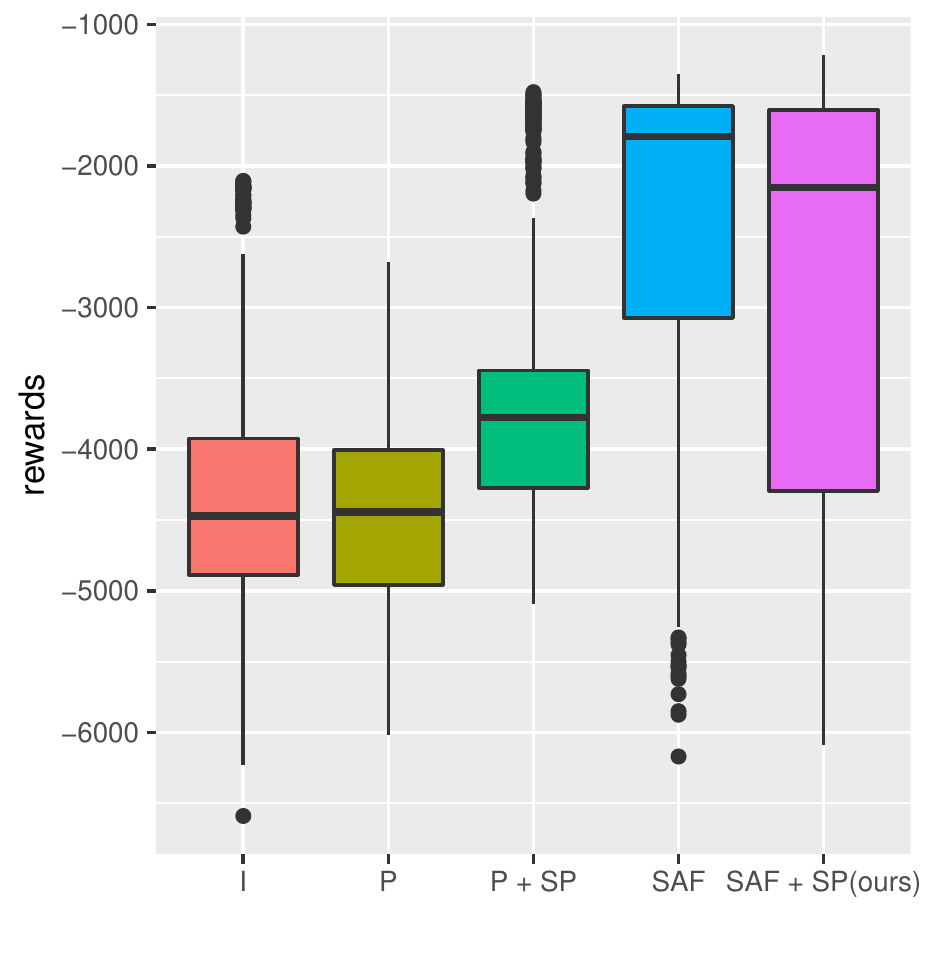}
    }\\
    \subfigure[15 agents]{
    \includegraphics[width=0.3\linewidth]{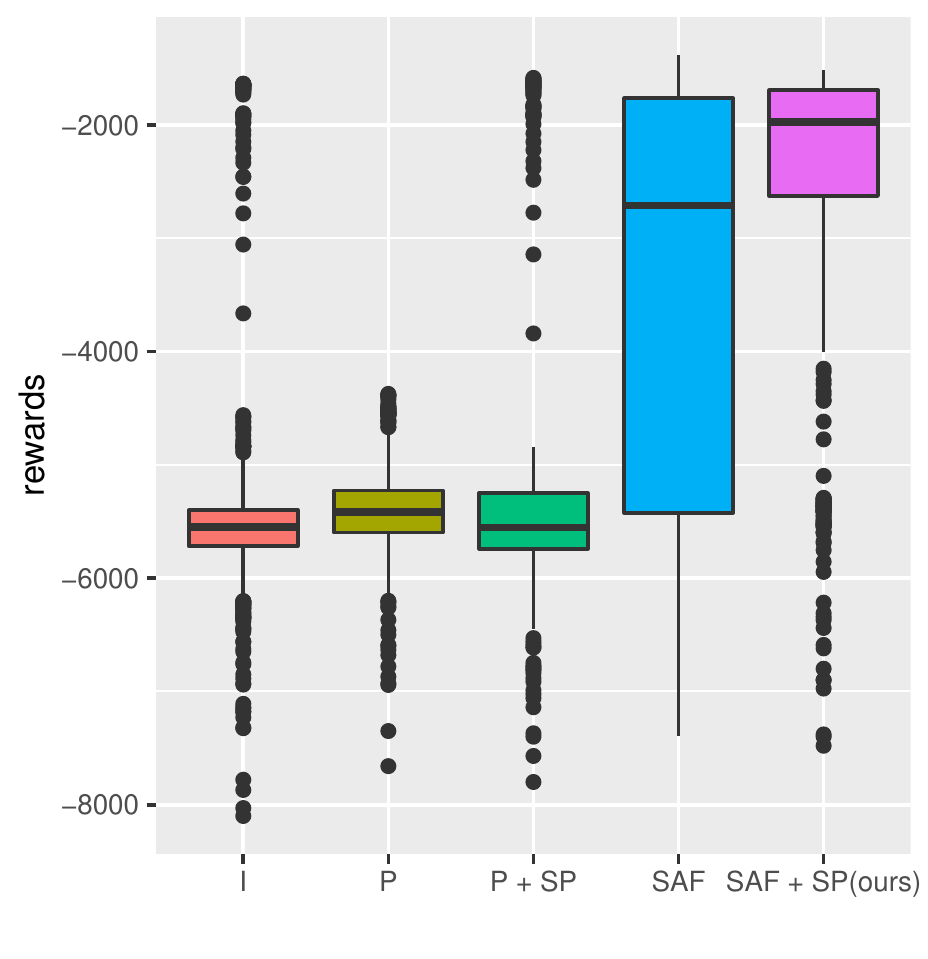}
    }
    \subfigure[30 agents]{
    \includegraphics[width=0.3\linewidth]{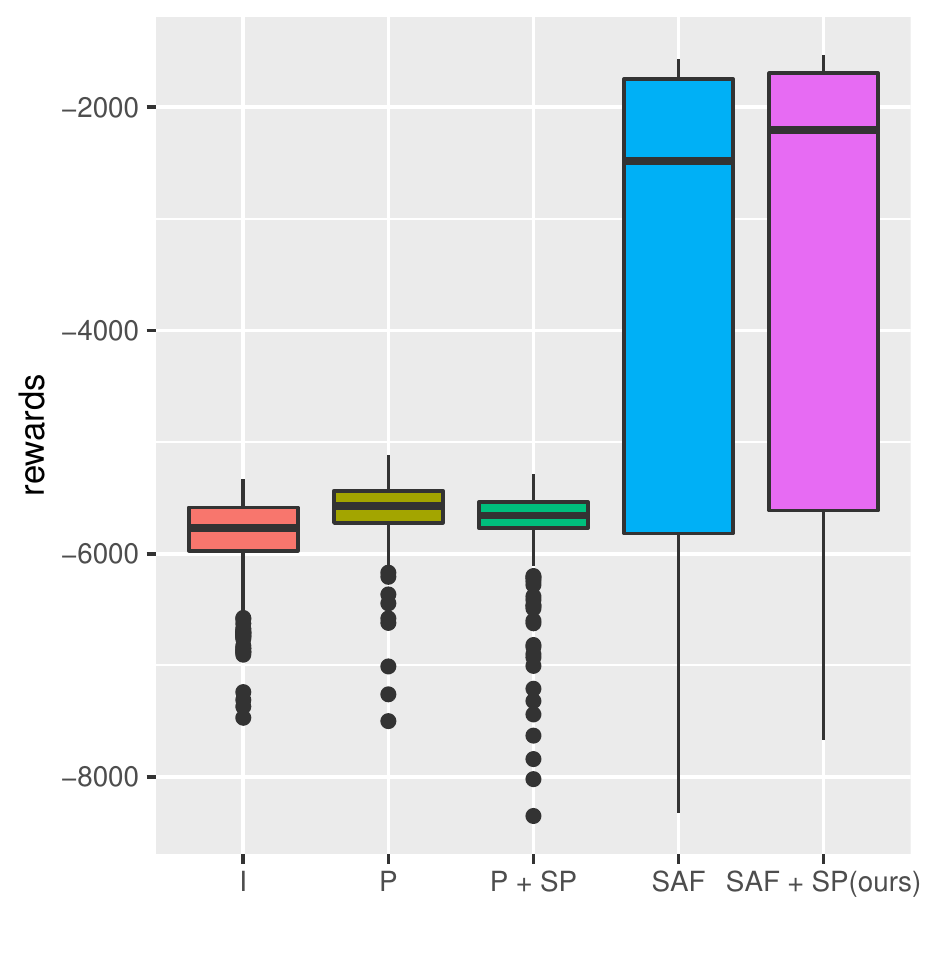}
    }

    % {
    %\includegraphics[width=0.8\linewidth]{figures/legendScale.png}
    %}
    \caption{\textsc{\saf}, shared policy pool and maximizing agent's independence in cooperative MARL bring significant improvement in performance when there are more cooperative agents in the same environment in the training and evaluation process. (a) 2 cooperative agents  (b) 5 cooperative agents (c) 15 cooperative agents (d) 30 cooperative agents. Our method (\emph{\saf} + SP) performs better when there is a  larger number of agents in the environment .All experiments in this figure were conducted in GhostRun environment}
    \label{fig:Scale}
\end{figure*}

%\subsection{Ablation Studies}

\begin{figure*}[ht!]
    \centering
    \subfigure[Communication Methods]{
    \includegraphics[width=0.2\linewidth]{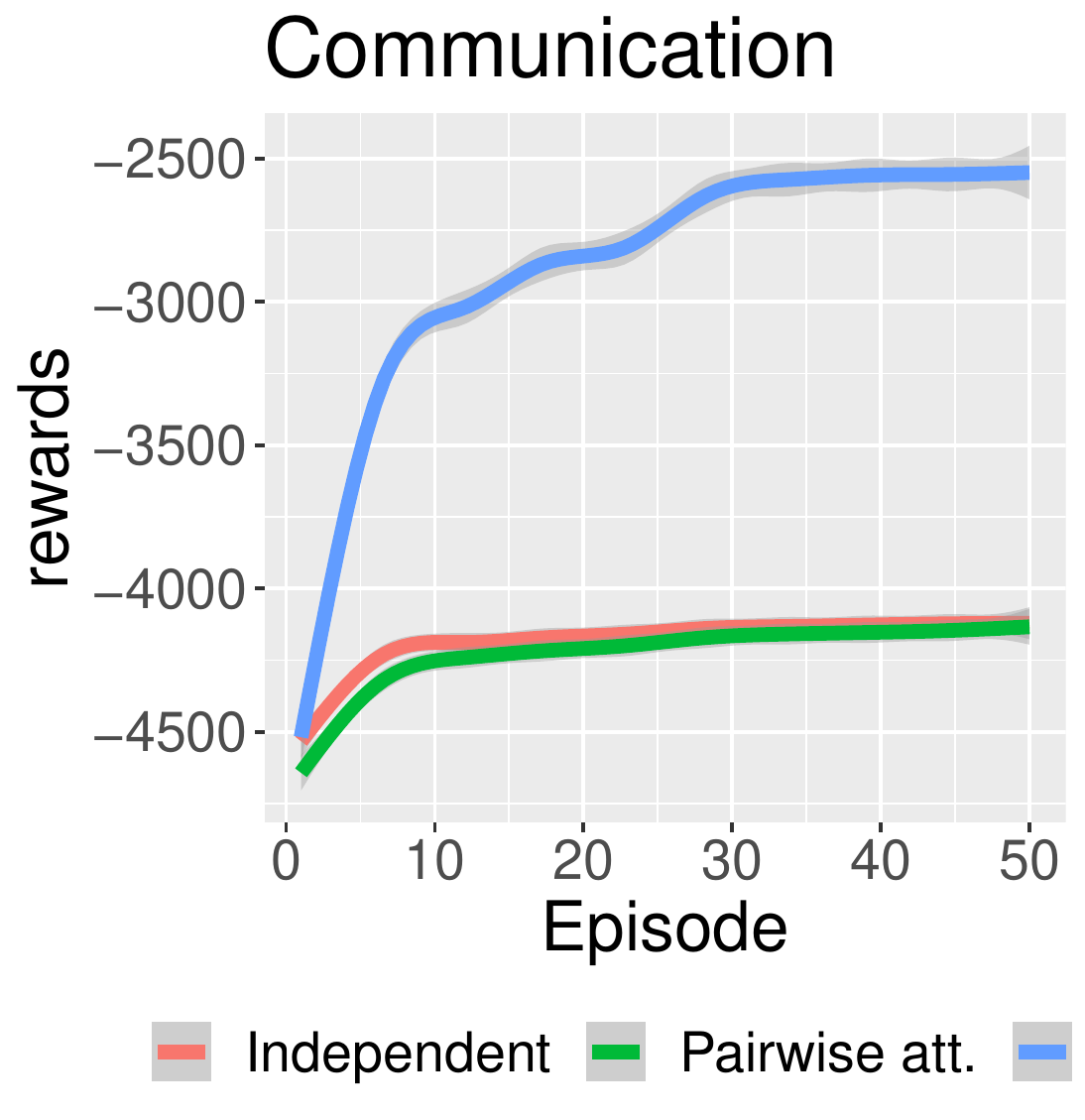}
    }
    \subfigure[Policy Selection]{
    \includegraphics[width=0.2\linewidth]{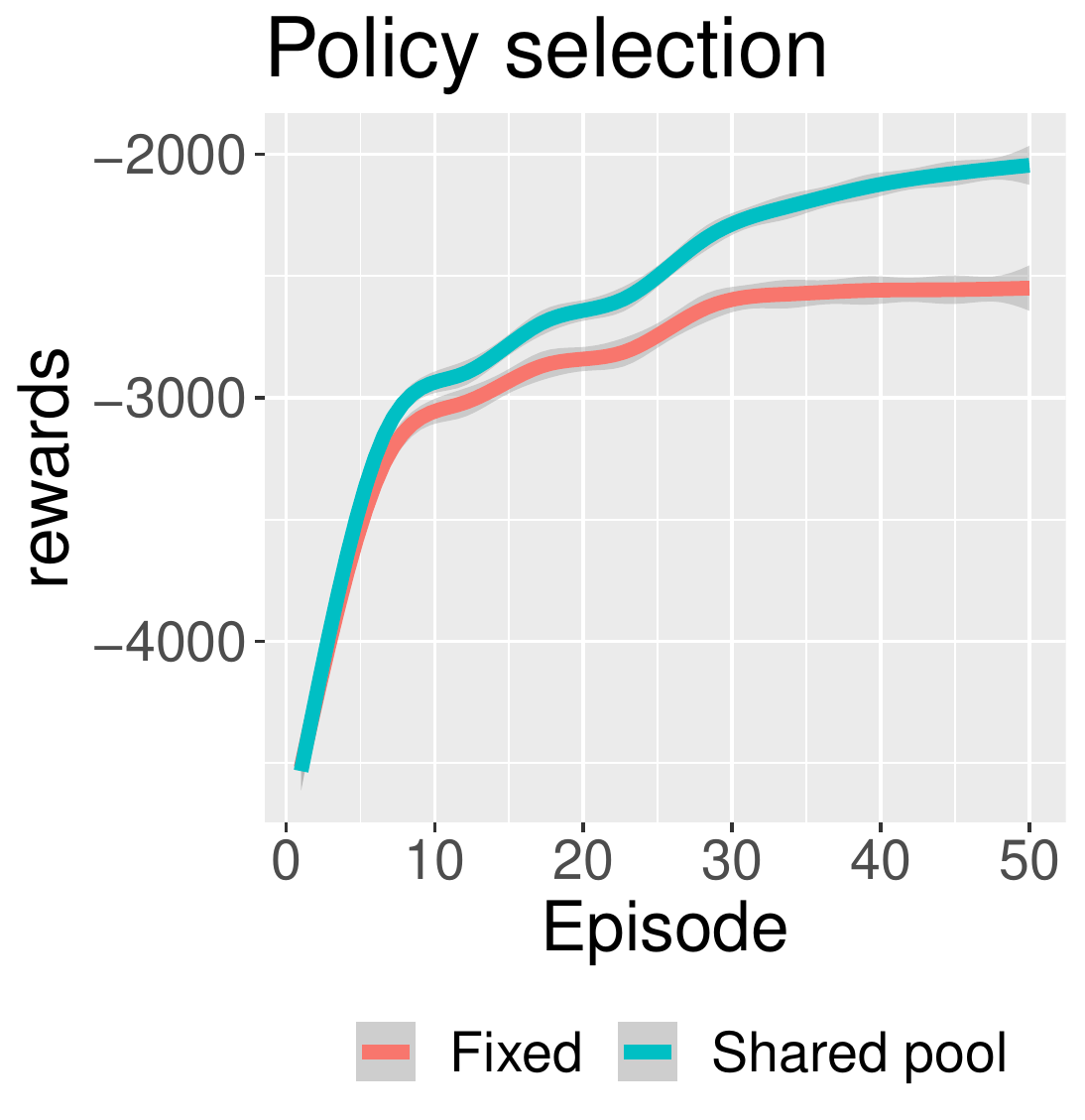}
    }
    \subfigure[Varying beta]{
    \includegraphics[width=0.2\linewidth]{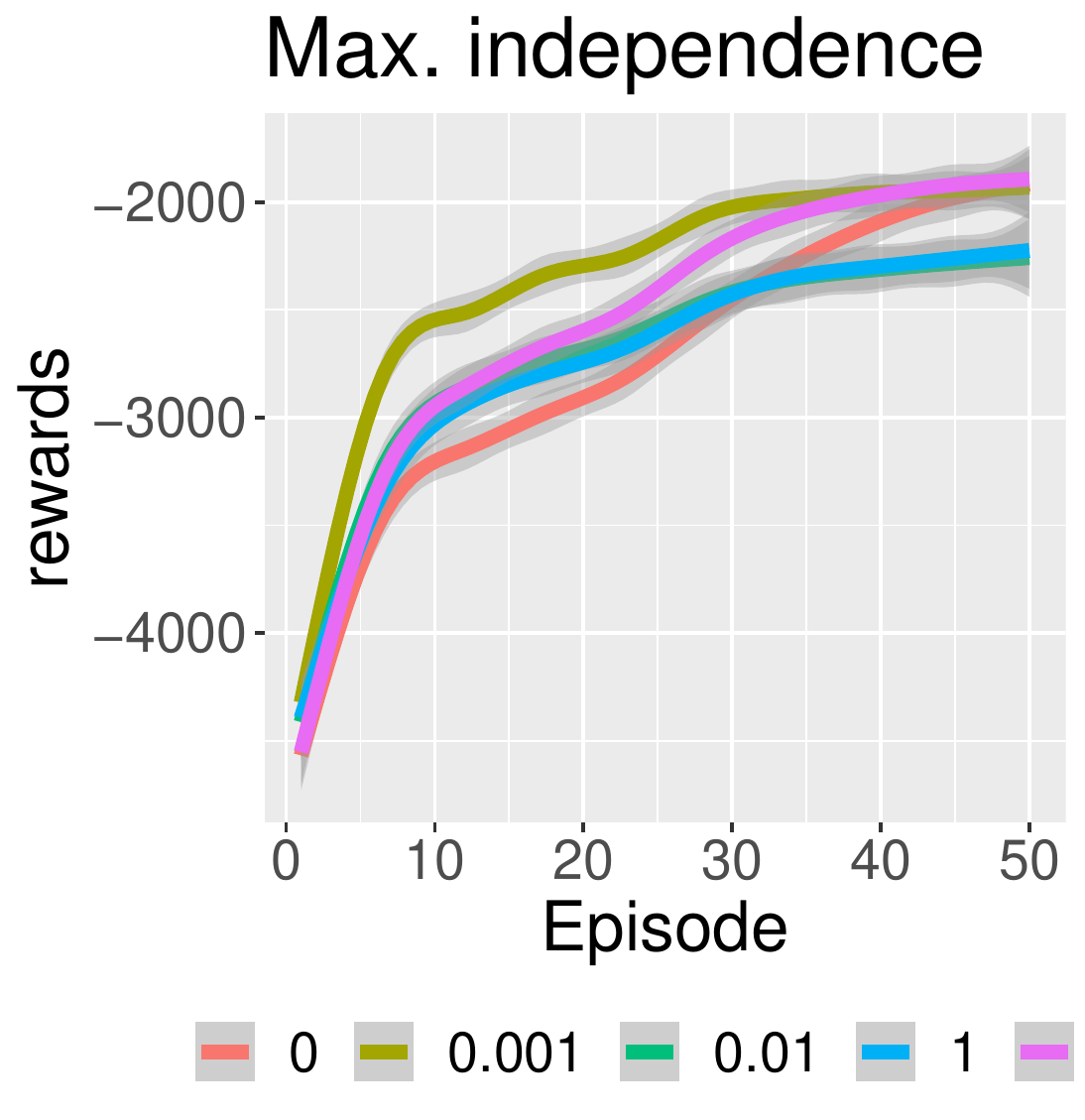}
    }
    \subfigure[Maximizing Independence]{
    \includegraphics[width=0.2\linewidth]{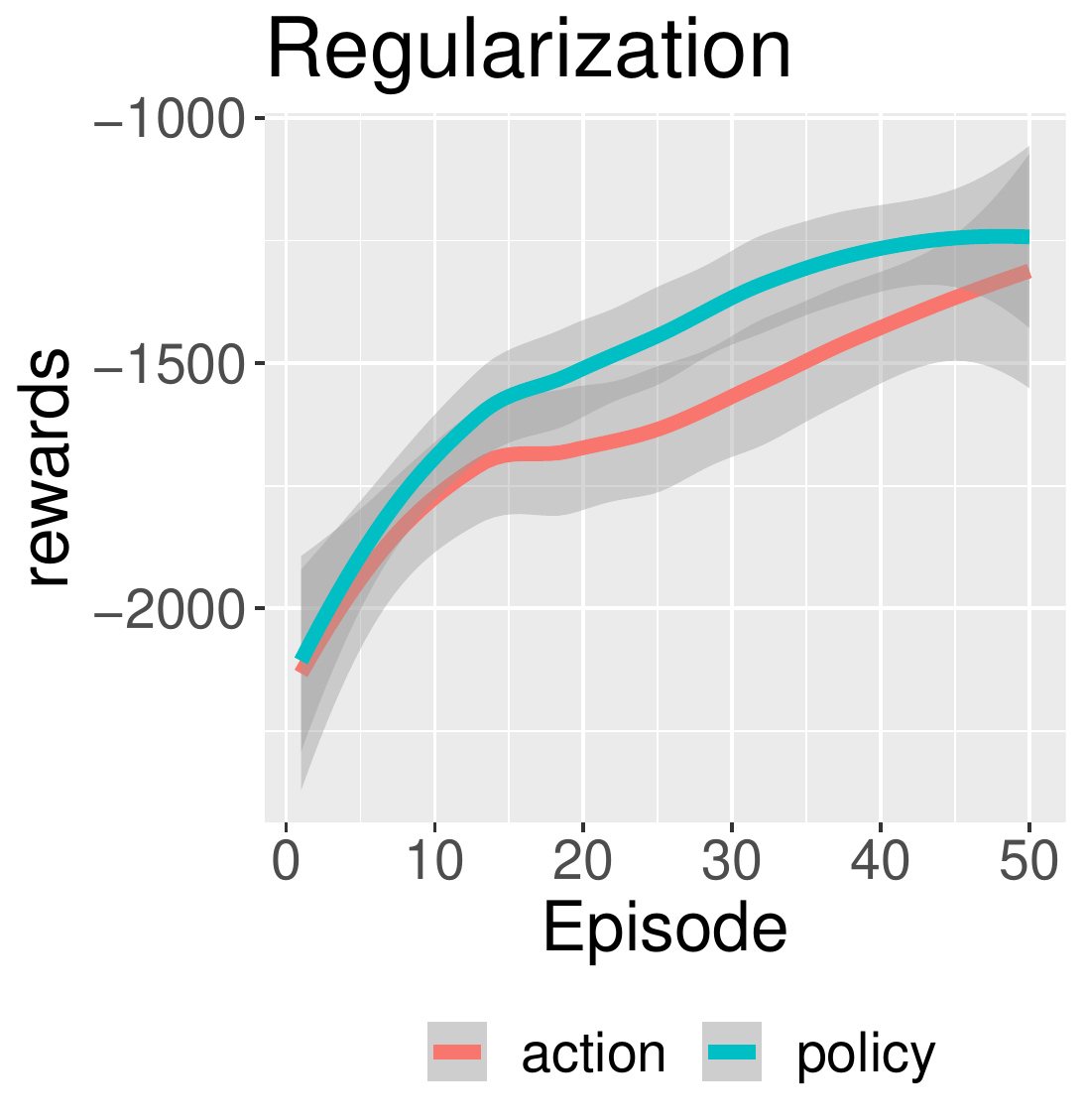}
    }
    \caption{\emph{\saf}, shared policy pool and maximizing agent's independence in cooperative MARL. (a) Comparison between different communication setups (b) Comparison between different policy selection strategies (c) Comparison between coefficients for KL Loss. (d) Comparison between penalizing policy selection and penalizing action selection}
    \label{fig:GhostRun}
\end{figure*}

%We perform the following ablation studies and analyze the results obtained for the Ghost Run Environment to determine the importance and contribution of each component towards the final performance of the approach: 
\vspace{-3mm}
\paragraph{Stateful and Active communication Facilitator (\emph{\saf}) improves communication in MARL.}
We investigate the use of \emph{\saf} for coordinating among different agents in the Ghost Run environment by comparing different communication schemes. (a). We compare three communication strategies [\textsc{I}], [\textsc{P}] and using \emph{\saf}. We find that the proposed \emph{\saf} based approach outperforms the other two settings and shows an improvement of about \textbf{45\%} over the other two baselines as shown in figure \ref{fig:GhostRun}(a). This shows that introducing the \emph{\saf} helps improve inter-agent communication. The \emph{\saf} acts as a specialist agent / communication bottleneck for the messages sent by all agents. It filters out only the relevant information for providing the relevant \textit{context} to each agent making it more robust, which may not be possible in pairwise communication i.e., where every pair of agents communicate with each other. 

We also compare [\emph{\saf}+SP] against a graph-based communication baseline \citet{wang2021tom2c} that we simply name \emph{\textsc{Graph}}, on the MSTC environment. In \emph{\textsc{Graph}}, the agents generate a message from their observation and communication is done through a fully-connected graph whose nodes are the updated messages. These are used top optimize the agents' policies. We show the test reward  in Table \ref{tom-saf}.
% left panel.In GhostRun environment five cooperative agents are trained using PPO algorithm,  we observe that with a SAF that integrates information from and makes state available to agents,the swarm show significant improvement performance agents communicating with each other via pairwise attention and agents act independently without communication.

\begin{figure*}[t]
    \centering
    \includegraphics[width=0.26\linewidth]{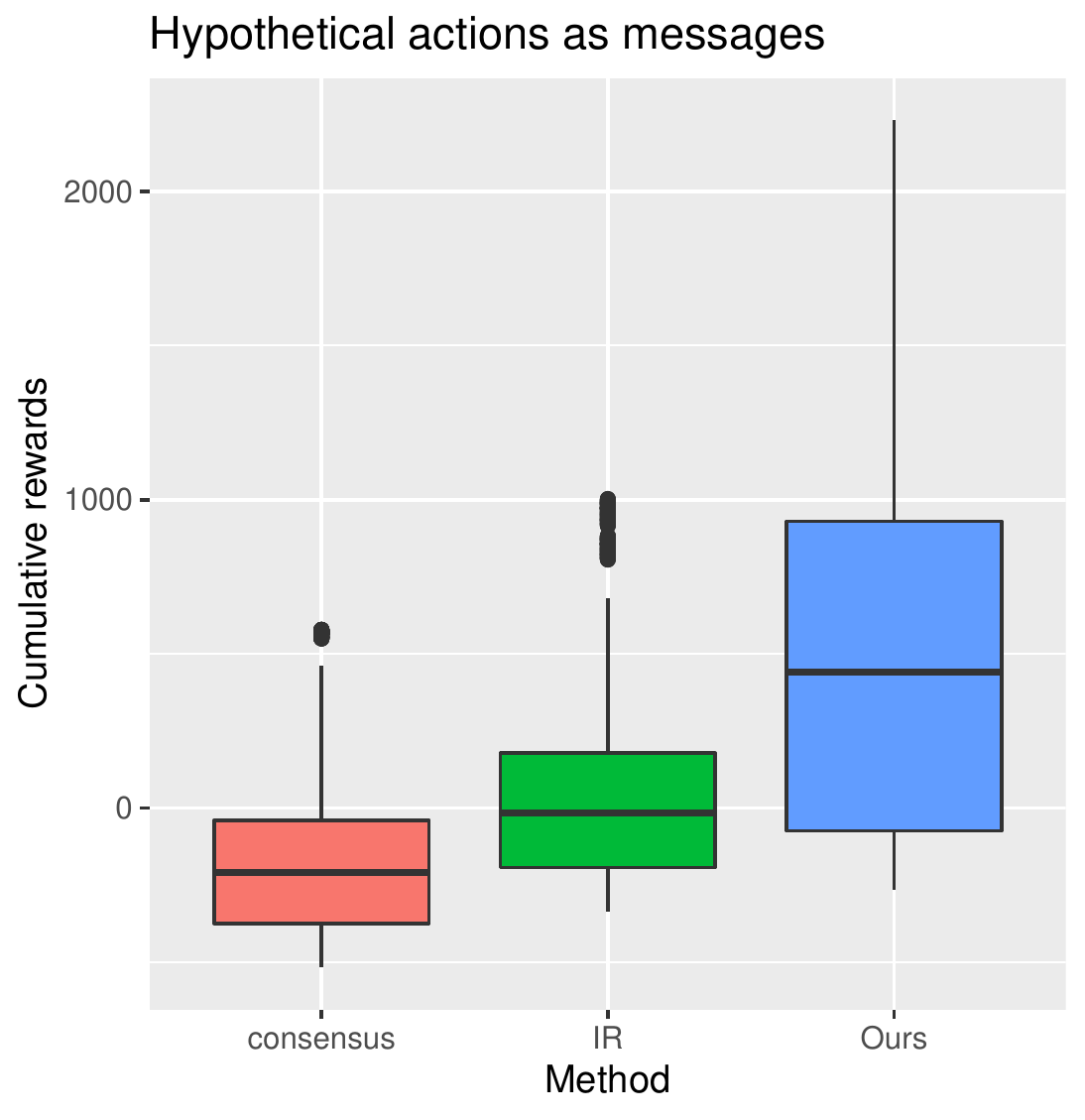}
    \includegraphics[width=0.26\linewidth]{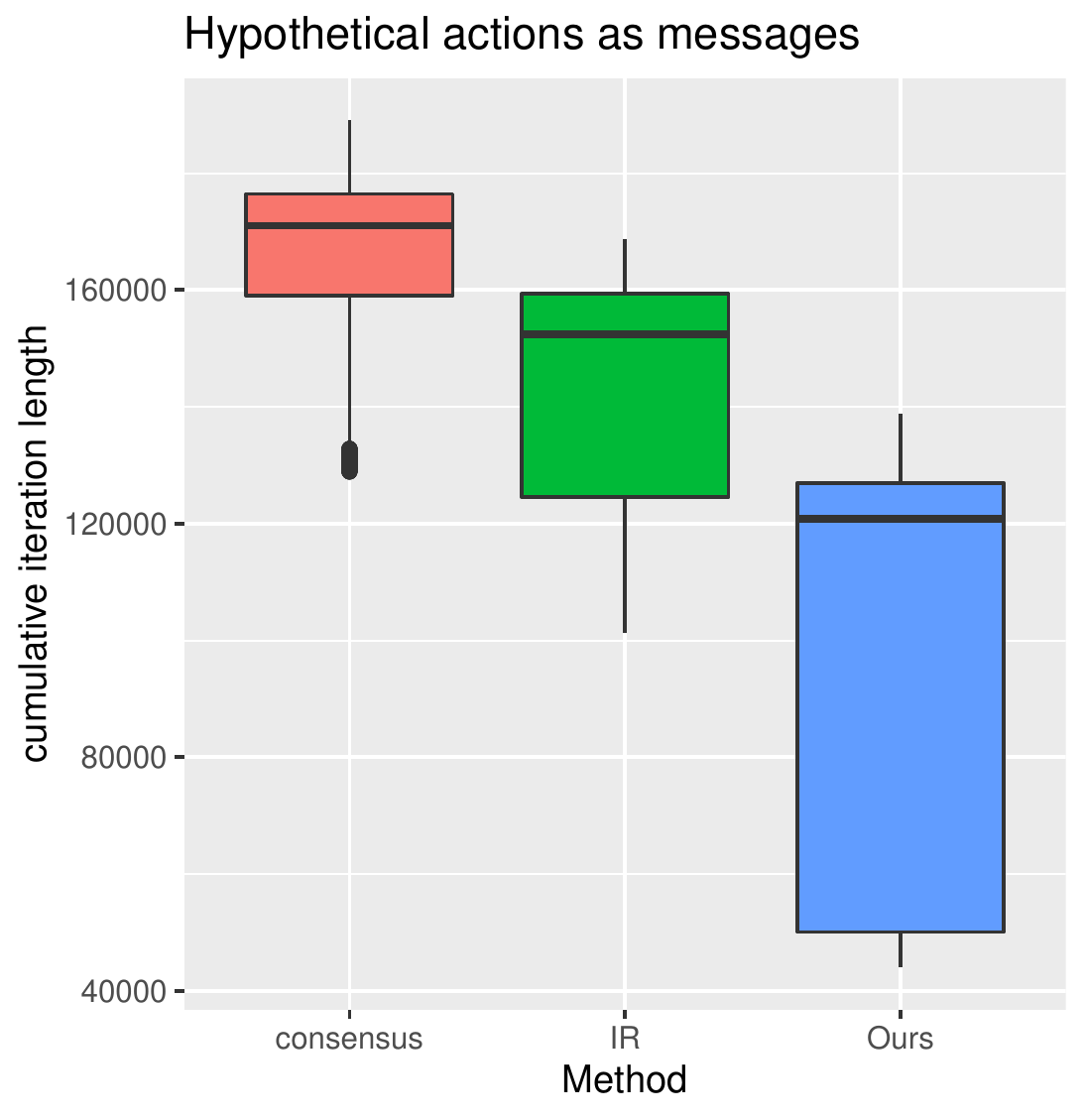}
    \caption{\emph{\saf}, shared policy pool and maximising agent's independence in cooperative MARL with agents communicating hypothetical actions as messages in PistonBall environment. Our method has higher cumulative rewards while decreasing the average episode's length compared to the baselines.}
    \label{fig:HypotheticalActions}
\end{figure*}

\begin{table}[]
\caption{Average reward on the test setting in the MSTC environment for the [\emph{\saf}+SP] architecture and the \emph{\textsc{Graph}} model}
\label{tom-saf}
\vskip 0in %%%%%%%%%%%%%%%%%%%%%%%%%%%
\begin{center}
\begin{small}
\begin{sc}
\begin{tabular}{lll}
\toprule
 & [\emph{\saf}+SP] & \emph{\textsc{Graph}} \\
\midrule
Expected Return   & $71.26 (\pm 5.81)$ &   $61.22 (\pm 8.15) $    \\
\bottomrule
\end{tabular}
\end{sc}
\end{small}
\end{center}
\vskip -0.1in
\end{table}

\vspace{-3mm}
\paragraph{Policy switching using shared policy pool improves performance.} We investigate the significance of using policy switching using a shared pool of policies for each agent. We experiment with two settings: \textsc{\emph{\saf}} and \textsc{\emph{\saf} + SP}.  The inter-agent communication in both settings is facilitated via the use of \emph{\saf}. The results are shown in figure \ref{fig:GhostRun}(b). Using a shared pool of policies with dynamic selection shows a relative improvement of \textbf{25\%} over the fixed policy setting.  This shows that the mixture of policies helps to transfer knowledge about different behaviors across different agents.

%restricts the the \emph{\saf} acting as a centralized controller and provides generalization advantages since each policy in the pool can acts a a specialist \emph{operator} for different kinds of states that an agent might be in. The fixed policy case can be essentially considered as a central controller based setting with inter-agent communication mediated by the \emph{\saf}. Using a fixed policy doesn't allow decomposition of the learning problem which can be done in the case of using a pool of policies and hence shows lower performance than that. %performance of agents with or without access to policy pool and policy switching are compared.The five agents have access to a pool of three different policies between which they can switch to for action prediction at each time step during training and execution. In the experiment setting, where agents do not conduct policy switching all agents share a single policy network. Our results shows reasonable improvement is achieved by using policy switching.

%To  reduce inter-agent dependence, we propose to minimize the dependence of each agent's policy on the information that is available from the \emph{\saf} in the form of its state, quantified by the conditional mutual information $I\!\left(A;M \mid S\right)$.

\vspace{-3mm}
\paragraph{Minimizing dependence on \emph{\saf}.} Here, we investigate the effect of minimizing the dependence of each agent's policy on the information that is available from the \emph{\saf} in the form of its state, quantified by the conditional mutual information $I\!\left(Z;M \mid S\right)$, where $Z$ is the agent's policy choice. One can directly optimize it  as discussed in eq.\ref{eq:information_bottleneck}. We also experimented with different values of the coefficients ($\beta$) of the KL based regularization loss. Figure \ref{fig:GhostRun}(c) shows that minimizing the dependence of each agent on the information from \emph{\saf} helps improve convergence. 

We also compare the effect of directly optimizing the conditional mutual information $I\!\left(A;M \mid S\right)$ i.e., where the information from \emph{\saf} is used to also select actions (as compared when information from \emph{\saf} is only used to select the policy) and the agent is optimized to minimize dependence on information from \emph{\saf}. We call the method where we optimize the mutual information directly on actions as \textit{regularized action selection}, and the method where we optimize the upper bound on mutual information as \textit{regularized policy selection}. Figure \ref{fig:GhostRun}(d) compares the effect of regularizing action as compared to regularizing the policy selection, and shows that regularizing the policy selection (i.e., upper bound) achieves better results as compared to regularizing the action selection.

\paragraph{\textsc{SAF} is significantly helpful for MARL training with large number of agents.} Here we investigate how the proposed method scales with increasing the number of agents. Learning a stateful, active facilitator \emph{\saf} which integrates information across agents should scale better as compared to using a communication protocol where every agent interacts with every other agent. To test this hypothesis, we compare the performance of five different approaches: \emph{\saf}, [\emph{\saf}+\textsc{SP}], [\textsc{P}], [\textsc{P} + \textsc{SP}], [\textsc{I}] by varying the number of agents. Figure \ref{fig:Scale} shows the result of these different methods by varying the number of agents. As the number of agents increases, the relative difference between the proposed method and the different baselines ([\textsc{I}], [\textsc{P}], [\textsc{P}+\textsc{SP}]) increase significantly. 

%\vspace{-2mm}
\paragraph{Flexible plug-in tool for many different MARL settings.} In the previous sections, we showed that \emph{\saf}, shared policy, and maximizing independence boost performance on cooperative MARL tasks in which agents communicate messages generated as a function of the observed environment. In this section, we investigate the possibility of applying the proposed method on cooperative MARL algorithms in which agents iteratively communicate \textit{hypothetical actions} with each other before taking actions. The performance of the proposed method is compared with two recent approaches that communicate hypothetical actions, namely consensus update and iterative reasoning (IR) \cite{zhang2018fully}. Consensus update conducts graph-based local communication and IR communicate hypothetical actions among agents such that an agent's policy is conditional on its teammates. Our experimental results show that our method achieves higher cumulative rewards in the PistonBall environment while doing so in a lower number of iterations, both of which indicate better performance (see Figure \ref{fig:HypotheticalActions}). This suggests the potential of the proposed method as a flexible plug-in tool in many MARL settings. 

%\vspace{-3mm}
% \paragraph{Zero shot generalization by increasing the number of agents.}
% Here we evaluate the performance of the proposed approach when during evaluating time, we increase the number of agents as compared to the number of agents seen during training time. 

\section{Future Work}
Here, we  introduced three key ideas that operate synergistically. First, communication among agents is conducted via an intelligent channel, \saf. Second, each agent is incentivized to act independently and avoid relying on \saf. Third, agents
operate according to a mixture-of-policies, where \saf\ provides the signal to select the policy. Through extensive experiments we show the utility of all the different components. We also show that the proposed method scales better on increasing the number of the agents and achieves higher returns as compared to various different baselines. Future work may involve scaling the proposed method to more complex multi-agent problems like Starcraft \citet{vinyals2019grandmaster} and Google Research Football \citet{gfootball}.

\bibliography{ICML_ref}

\begin{thebibliography}{}

\bibitem[Baars, 1988]{Baars1988-BAAACT}
Baars, B.~J. (1988).
\newblock {\em A Cognitive Theory of Consciousness}.
\newblock Cambridge University Press.

\bibitem[Bacon et~al., 2017]{Bacon2017The}
Bacon, P.-L., Harb, J., and Precup, D. (2017).
\newblock The option-critic architecture.
\newblock {\em Proceedings of the AAAI Conference on Artificial Intelligence},
  31(1).

\bibitem[Bagaria and Konidaris, 2020]{Bagaria2020Option}
Bagaria, A. and Konidaris, G. (2020).
\newblock Option discovery using deep skill chaining.
\newblock In {\em International Conference on Learning Representations}.

\bibitem[Bahdanau et~al., 2014]{bahdanau2014neural}
Bahdanau, D., Cho, K., and Bengio, Y. (2014).
\newblock Neural machine translation by jointly learning to align and
  translate.
\newblock {\em arXiv preprint arXiv:1409.0473}.

\bibitem[Battaglia et~al., 2018]{battaglia2018relational}
Battaglia, P.~W., Hamrick, J.~B., Bapst, V., Sanchez-Gonzalez, A., Zambaldi,
  V., Malinowski, M., Tacchetti, A., Raposo, D., Santoro, A., Faulkner, R.,
  et~al. (2018).
\newblock Relational inductive biases, deep learning, and graph networks.
\newblock {\em arXiv preprint arXiv:1806.01261}.

\bibitem[Bronstein et~al., 2017]{bronstein2017geometric}
Bronstein, M.~M., Bruna, J., LeCun, Y., Szlam, A., and Vandergheynst, P.
  (2017).
\newblock Geometric deep learning: going beyond euclidean data.
\newblock {\em IEEE Signal Processing Magazine}, 34(4):18--42.

\bibitem[Busoniu et~al., 2008]{busoniu2008comprehensive}
Busoniu, L., Babuska, R., and De~Schutter, B. (2008).
\newblock A comprehensive survey of multiagent reinforcement learning.
\newblock {\em IEEE Transactions on Systems, Man, and Cybernetics, Part C
  (Applications and Reviews)}, 38(2):156--172.

\bibitem[Chakravorty et~al., 2020]{Chakravorty2020Option}
Chakravorty, J., Ward, P.~N., Roy, J., Chevalier-Boisvert, M., Basu, S., Lupu,
  A., and Precup, D. (2020).
\newblock Option-critic in cooperative multi-agent systems.
\newblock In {\em Proceedings of the 19th International Conference on
  Autonomous Agents and MultiAgent Systems}, AAMAS '20, page 1792–1794,
  Richland, SC. International Foundation for Autonomous Agents and Multiagent
  Systems.

\bibitem[Das et~al., 2019]{das2019tarmac}
Das, A., Gervet, T., Romoff, J., Batra, D., Parikh, D., Rabbat, M., and Pineau,
  J. (2019).
\newblock Tarmac: Targeted multi-agent communication.
\newblock In {\em International Conference on Machine Learning}, pages
  1538--1546. PMLR.

\bibitem[Dehaene et~al., 2017]{Dehaene-et-al-2017}
Dehaene, S., Lau, H., and Kouider, S. (2017).
\newblock What is consciousness, and could machines have it?
\newblock {\em Science}, 358(6362):486--492.

\bibitem[Foerster et~al., 2016]{foerster2016learning}
Foerster, J.~N., Assael, Y.~M., De~Freitas, N., and Whiteson, S. (2016).
\newblock Learning to communicate with deep multi-agent reinforcement learning.
\newblock {\em arXiv preprint arXiv:1605.06676}.

\bibitem[Foerster et~al., 2018a]{IAC}
Foerster, J.~N., Farquhar, G., Afouras, T., Nardelli, N., and Whiteson, S.
  (2018a).
\newblock Counterfactual multi-agent policy gradients.
\newblock In {\em AAAI}.

\bibitem[Foerster et~al., 2018b]{Foerster2018CounterfactualMP}
Foerster, J.~N., Farquhar, G., Afouras, T., Nardelli, N., and Whiteson, S.
  (2018b).
\newblock Counterfactual multi-agent policy gradients.
\newblock In {\em AAAI}.

\bibitem[Galashov et~al., 2019]{galashov2019information}
Galashov, A., Jayakumar, S.~M., Hasenclever, L., Tirumala, D., Schwarz, J.,
  Desjardins, G., Czarnecki, W.~M., Teh, Y.~W., Pascanu, R., and Heess, N.
  (2019).
\newblock Information asymmetry in kl-regularized rl.
\newblock {\em arXiv preprint arXiv:1905.01240}.

\bibitem[Goyal et~al., 2021a]{goyal2021neural}
Goyal, A., Didolkar, A., Ke, N.~R., Blundell, C., Beaudoin, P., Heess, N.,
  Mozer, M., and Bengio, Y. (2021a).
\newblock Neural production systems.
\newblock {\em arXiv preprint arXiv:2103.01937}.

\bibitem[Goyal et~al., 2021b]{goyal2021coordination}
Goyal, A., Didolkar, A., Lamb, A., Badola, K., Ke, N.~R., Rahaman, N., Binas,
  J., Blundell, C., Mozer, M., and Bengio, Y. (2021b).
\newblock Coordination among neural modules through a shared global workspace.
\newblock {\em arXiv preprint arXiv:2103.01197}.

\bibitem[Goyal et~al., 2019]{goyal2019infobot}
Goyal, A., Islam, R., Strouse, D., Ahmed, Z., Botvinick, M., Larochelle, H.,
  Levine, S., and Bengio, Y. (2019).
\newblock Infobot: Transfer and exploration via the information bottleneck.
\newblock {\em arXiv preprint arXiv:1901.10902}.

\bibitem[Goyal et~al., 2021c]{goyal2021scoff}
Goyal, A., Lamb, A., Gampa, P., Beaudoin, P., Levine, S., Blundell, C., Bengio,
  Y., and Mozer, M.~C. (2021c).
\newblock {Object files and schemata: Factorizing declarative and procedural
  knowledge in dynamical systems}.
\newblock In {\em International Conference on Learning Representations}.

\bibitem[Gupta et~al., 2017]{gupta2017cooperative}
Gupta, J.~K., Egorov, M., and Kochenderfer, M. (2017).
\newblock Cooperative multi-agent control using deep reinforcement learning.
\newblock In {\em International Conference on Autonomous Agents and Multiagent
  Systems}, pages 66--83. Springer.

\bibitem[Han et~al., 2019]{Han2019Multi}
Han, D., Boehmer, W., Wooldridge, M., and Rogers, A. (2019).
\newblock Multi-agent hierarchical reinforcement learning with dynamic
  termination.
\newblock AAMAS '19, page 2006–2008, Richland, SC. International Foundation
  for Autonomous Agents and Multiagent Systems.

\bibitem[Jang et~al., 2016]{jang2016categorical}
Jang, E., Gu, S., and Poole, B. (2016).
\newblock Categorical reparameterization with gumbel-softmax.
\newblock {\em arXiv preprint arXiv:1611.01144}.

\bibitem[Jiang and Lu, 2018]{Jiang2018LearningAC}
Jiang, J. and Lu, Z. (2018).
\newblock Learning attentional communication for multi-agent cooperation.
\newblock In {\em NeurIPS}.

\bibitem[Jiang, 2019]{shuo2019maenvs}
Jiang, S. (2019).
\newblock Multi agent reinforcement learning environments compilation.

\bibitem[Jiang and Amato, 2021]{jiang2021multi}
Jiang, S. and Amato, C. (2021).
\newblock Multi-agent reinforcement learning with directed exploration and
  selective memory reuse.
\newblock In {\em Proceedings of the 36th Annual ACM Symposium on Applied
  Computing}, pages 777--784.

\bibitem[Khetarpal et~al., 2020]{Khetarpal2020OptionsOI}
Khetarpal, K., Klissarov, M., Chevalier-Boisvert, M., Bacon, P.-L., and Precup,
  D. (2020).
\newblock Options of interest: Temporal abstraction with interest functions.
\newblock In {\em AAAI}.

\bibitem[Kim et~al., 2020]{kim2020communication}
Kim, W., Park, J., and Sung, Y. (2020).
\newblock Communication in multi-agent reinforcement learning: Intention
  sharing.
\newblock In {\em International Conference on Learning Representations}.

\bibitem[Kipf et~al., 2018]{kipf2018neural}
Kipf, T., Fetaya, E., Wang, K.-C., Welling, M., and Zemel, R. (2018).
\newblock Neural relational inference for interacting systems.
\newblock {\em arXiv preprint arXiv:1802.04687}.

\bibitem[Klissarov et~al., 2017]{Klissarov2017Learning}
Klissarov, M., Bacon, P.-L., Harb, J., and Precup, D. (2017).
\newblock Learning options end-to-end for continuous action tasks.

\bibitem[Kurach et~al., 2019]{gfootball}
Kurach, K., Raichuk, A., Stańczyk, P., Zając, M., Bachem, O., Espeholt, L.,
  Riquelme, C., Vincent, D., Michalski, M., Bousquet, O., and Gelly, S. (2019).
\newblock Google research football: A novel reinforcement learning environment.

\bibitem[Li et~al., 2021]{li2021celebrating}
Li, C., Wu, C., Wang, T., Yang, J., Zhao, Q., and Zhang, C. (2021).
\newblock Celebrating diversity in shared multi-agent reinforcement learning.
\newblock {\em arXiv preprint arXiv:2106.02195}.

\bibitem[Liu et~al., 2021]{DianboLiu2021DiscreteValuedNC}
Liu, D., Lamb, A., Kawaguchi, K., Goyal, A., Sun, C., Mozer, M.~C., and Bengio,
  Y. (2021).
\newblock Discrete-valued neural communication.
\newblock {\em ArXiv}, abs/2107.02367.

\bibitem[Lowe et~al., 2017]{lowe2017multi}
Lowe, R., Wu, Y., Tamar, A., Harb, J., Abbeel, P., and Mordatch, I. (2017).
\newblock Multi-agent actor-critic for mixed cooperative-competitive
  environments.
\newblock {\em arXiv preprint arXiv:1706.02275}.

\bibitem[Mahajan et~al., 2019]{Mahajan2019MAVENMV}
Mahajan, A., Rashid, T., Samvelyan, M., and Whiteson, S. (2019).
\newblock Maven: Multi-agent variational exploration.
\newblock In {\em NeurIPS}.

\bibitem[Nachum et~al., 2018]{Nachum2018Data}
Nachum, O., Gu, S., Lee, H., and Levine, S. (2018).
\newblock Data-efficient hierarchical reinforcement learning.
\newblock In {\em Proceedings of the 32nd International Conference on Neural
  Information Processing Systems}, NIPS'18, page 3307–3317. Curran Associates
  Inc.

\bibitem[Oord et~al., 2017]{oord2017neural}
Oord, A. v.~d., Vinyals, O., and Kavukcuoglu, K. (2017).
\newblock Neural discrete representation learning.
\newblock {\em arXiv preprint arXiv:1711.00937}.

\bibitem[Pateria et~al., 2021]{Pateria2021Hierarchical}
Pateria, S., Subagdja, B., Tan, A.-h., and Quek, C. (2021).
\newblock Hierarchical reinforcement learning: A comprehensive survey.
\newblock {\em ACM Comput. Surv.}, 54(5).

\bibitem[Rashid et~al., 2018a]{qmix}
Rashid, T., Samvelyan, M., Schroeder, C., Farquhar, G., Foerster, J., and
  Whiteson, S. (2018a).
\newblock Qmix: Monotonic value function factorisation for deep multi-agent
  reinforcement learning.
\newblock In {\em International Conference on Machine Learning}, pages
  4295--4304. PMLR.

\bibitem[Rashid et~al., 2018b]{Rashid2018QMIXMV}
Rashid, T., Samvelyan, M., Witt, C. S.~D., Farquhar, G., Foerster, J.~N., and
  Whiteson, S. (2018b).
\newblock Qmix: Monotonic value function factorisation for deep multi-agent
  reinforcement learning.
\newblock {\em ArXiv}, abs/1803.11485.

\bibitem[Riemer et~al., 2018]{Riemer2018LearningAO}
Riemer, M., Liu, M., and Tesauro, G. (2018).
\newblock Learning abstract options.
\newblock {\em ArXiv}, abs/1810.11583.

\bibitem[Scarselli et~al., 2008]{scarselli2008graph}
Scarselli, F., Gori, M., Tsoi, A.~C., Hagenbuchner, M., and Monfardini, G.
  (2008).
\newblock The graph neural network model.
\newblock {\em IEEE Transactions on Neural Networks}, 20(1):61--80.

\bibitem[Shanahan, 2006]{shanahan2006cognitive}
Shanahan, M. (2006).
\newblock A cognitive architecture that combines internal simulation with a
  global workspace.
\newblock {\em Consciousness and cognition}, 15(2):433--449.

\bibitem[Singh et~al., 2018]{singh2018learning}
Singh, A., Jain, T., and Sukhbaatar, S. (2018).
\newblock Learning when to communicate at scale in multiagent cooperative and
  competitive tasks.
\newblock {\em arXiv preprint arXiv:1812.09755}.

\bibitem[Sukhbaatar et~al., 2016]{sukhbaatar2016learning}
Sukhbaatar, S., Fergus, R., et~al. (2016).
\newblock Learning multiagent communication with backpropagation.
\newblock {\em Advances in neural information processing systems},
  29:2244--2252.

\bibitem[Sukhbaatar et~al., 2015]{Sukhbaatar2015}
Sukhbaatar, S., szlam, a., Weston, J., and Fergus, R. (2015).
\newblock End-to-end memory networks.
\newblock In Cortes, C., Lawrence, N.~D., Lee, D.~D., Sugiyama, M., and
  Garnett, R., editors, {\em Advances in Neural Information Processing Systems
  28}, pages 2440--2448. Curran Associates, Inc.

\bibitem[Sunehag et~al., 2018a]{vdn}
Sunehag, P., Lever, G., Gruslys, A., Czarnecki, W.~M., Zambaldi, V., Jaderberg,
  M., Lanctot, M., Sonnerat, N., Leibo, J.~Z., Tuyls, K., and Graepel, T.
  (2018a).
\newblock Value-decomposition networks for cooperative multi-agent learning
  based on team reward.
\newblock In {\em Proceedings of the 17th International Conference on
  Autonomous Agents and MultiAgent Systems}, AAMAS '18, page 2085–2087,
  Richland, SC. International Foundation for Autonomous Agents and Multiagent
  Systems.

\bibitem[Sunehag et~al., 2018b]{Sunehag2018ValueDecompositionNF}
Sunehag, P., Lever, G., Gruslys, A., Czarnecki, W.~M., Zambaldi, V.~F.,
  Jaderberg, M., Lanctot, M., Sonnerat, N., Leibo, J.~Z., Tuyls, K., and
  Graepel, T. (2018b).
\newblock Value-decomposition networks for cooperative multi-agent learning.
\newblock {\em ArXiv}, abs/1706.05296.

\bibitem[Sutton et~al., 1999]{Sutton1999Between}
Sutton, R.~S., Precup, D., and Singh, S. (1999).
\newblock Between mdps and semi-mdps: A framework for temporal abstraction in
  reinforcement learning.
\newblock {\em Artificial Intelligence}, 112(1):181--211.

\bibitem[Tan, 1993]{Tan1993MultiAgentRL}
Tan, M. (1993).
\newblock Multi-agent reinforcement learning: Independent versus cooperative
  agents.
\newblock In {\em ICML}.

\bibitem[Tan, 1997]{IQL}
Tan, M. (1997).
\newblock {\em Multi-Agent Reinforcement Learning: Independent vs. Cooperative
  Agents}, page 487–494.
\newblock Morgan Kaufmann Publishers Inc., San Francisco, CA, USA.

\bibitem[Teh et~al., 2017]{teh2017distral}
Teh, Y., Bapst, V., Czarnecki, W.~M., Quan, J., Kirkpatrick, J., Hadsell, R.,
  Heess, N., and Pascanu, R. (2017).
\newblock Distral: Robust multitask reinforcement learning.
\newblock In {\em Advances in Neural Information Processing Systems}, pages
  4496--4506.

\bibitem[Tirumala et~al., 2020]{tirumala2020behavior}
Tirumala, D., Galashov, A., Noh, H., Hasenclever, L., Pascanu, R., Schwarz, J.,
  Desjardins, G., Czarnecki, W.~M., Ahuja, A., Teh, Y.~W., et~al. (2020).
\newblock Behavior priors for efficient reinforcement learning.
\newblock {\em arXiv preprint arXiv:2010.14274}.

\bibitem[Vaswani et~al., 2017]{vaswani2017attention}
Vaswani, A., Shazeer, N., Parmar, N., Uszkoreit, J., Jones, L., Gomez, A.~N.,
  Kaiser, {\L}., and Polosukhin, I. (2017).
\newblock Attention is all you need.
\newblock In {\em Advances in neural information processing systems}, pages
  5998--6008.

\bibitem[Vinyals et~al., 2019]{vinyals2019grandmaster}
Vinyals, O., Babuschkin, I., Czarnecki, W.~M., Mathieu, M., Dudzik, A., Chung,
  J., Choi, D.~H., Powell, R., Ewalds, T., Georgiev, P., et~al. (2019).
\newblock Grandmaster level in starcraft ii using multi-agent reinforcement
  learning.
\newblock {\em Nature}, 575(7782):350--354.

\bibitem[Wang et~al., 2020]{wang2020qplex}
Wang, J., Ren, Z., Liu, T., Yu, Y., and Zhang, C. (2020).
\newblock Qplex: Duplex dueling multi-agent q-learning.
\newblock {\em arXiv preprint arXiv:2008.01062}.

\bibitem[Wang et~al., 2017]{wang2017nonlocal}
Wang, X., Girshick, R., Gupta, A., and He, K. (2017).
\newblock Non-local neural networks.

\bibitem[Wang et~al., 2021]{wang2021tom2c}
Wang, Y., Zhong, F., Xu, J., and Wang, Y. (2021).
\newblock Tom2c: Target-oriented multi-agent communication and cooperation with
  theory of mind.

\bibitem[Witt et~al., 2020a]{IPPO}
Witt, C. S.~D., Gupta, T., Makoviichuk, D., Makoviychuk, V., Torr, P. H.~S.,
  Sun, M., and Whiteson, S. (2020a).
\newblock Is independent learning all you need in the starcraft multi-agent
  challenge?
\newblock {\em ArXiv}, abs/2011.09533.

\bibitem[Witt et~al., 2020b]{Witt2020IsIL}
Witt, C. S.~D., Gupta, T., Makoviichuk, D., Makoviychuk, V., Torr, P. H.~S.,
  Sun, M., and Whiteson, S. (2020b).
\newblock Is independent learning all you need in the starcraft multi-agent
  challenge?
\newblock {\em ArXiv}, abs/2011.09533.

\bibitem[Yu et~al., 2021a]{yu2021surprising}
Yu, C., Velu, A., Vinitsky, E., Wang, Y., Bayen, A., and Wu, Y. (2021a).
\newblock The surprising effectiveness of ppo in cooperative, multi-agent
  games.
\newblock {\em arXiv preprint arXiv:2103.01955}.

\bibitem[Yu et~al., 2021b]{Yu2021TheSE}
Yu, C., Velu, A., Vinitsky, E., Wang, Y., Bayen, A.~M., and Wu, Y. (2021b).
\newblock The surprising effectiveness of mappo in cooperative, multi-agent
  games.
\newblock {\em ArXiv}, abs/2103.01955.

\bibitem[Zhang et~al., 2018]{zhang2018fully}
Zhang, K., Yang, Z., Liu, H., Zhang, T., and Basar, T. (2018).
\newblock Fully decentralized multi-agent reinforcement learning with networked
  agents.
\newblock In {\em International Conference on Machine Learning}, pages
  5872--5881. PMLR.

\end{thebibliography}

\bibliographystyle{apalike}

\appendix

\section{Appendix}

\subsection{Algorithm}
\label{appx:alg}

\begin{algorithm*}[h]
\caption{MARL with \emph{\saf}, Maximizing Agent Independence and Policy Switching}
   \label{alg:attention}
   %\begin{algorithmic}%[1]
   \footnotesize
   
    \For{$t \gets 1$ \KwTo $T$}  {
    \vspace{1mm}
    \textbf{Step 1: Each agent $i$ having state information $s_{i,t}$ (encoded partial observation), generates a message}. \\
    $\forall i\in\{1,\dots,N\}, m'_{i,t}=f_{\theta}(s_{i,t}) $\\
    $M'_t=(m'_{1,t},m'_{1,t},m'_{2,t}...,m'_{N,t})$

    \BlankLine

    \vspace{3mm}    
    \textbf{Step 2: \emph{\saf} integrates information from all agents} \\
    $\widetilde{\vQ} = F_t\widetilde{\vW}^{q}$\\
    $\vF_{t} \leftarrow \mathrm{softmax}\left(\frac{\widetilde{\vQ}(\vM'_{t} \widetilde{\vW}^{e})^\mathrm{T}}{\sqrt{d_e}}\right)\vM'_{t} \widetilde{\vW}^{v}$\\
    Self-attention over $F_t$ to update the \emph{\saf} state.
    \BlankLine
    
    \vspace{3mm}
    \textbf{Step 3: Information from \emph{\saf} is made available to each agent} \\
    $$
    \begin{cases}
    q^{s}_{i,t} &= W^{q}_{write}s_{i, t} \\
    \kappa &= (F_tW^e)^T \\
    \alpha &= \mathrm{softmax} \left( \frac{Q^s_t\kappa}{\sqrt{d_e}}\right) \\
    M_t &= \mathrm{softmax} \left( \frac{Q^s_t\kappa}{\sqrt{d_e}}\right)F_tW^v
    \end{cases}
    $$
   
   \BlankLine
    
    \vspace{3mm}
    \textbf{Step 4: Policy Selection from the pool} \\
    
    $$\forall i\in\{1,\dots,N\} 
    \begin{cases}
    q^{policy}_{i, t} &= f_{psel}(s_{i, t},  m_{i, t})\\
    \mathrm{index}_i &= \mathrm{GumbelSoftmax}\left(\frac{q^{policy}_{i, t} (K^{\Pi}_t)^T}{\sqrt{d_m}}\right)
    \end{cases}
    $$
    
    \vspace{3mm}
    \textbf{Step 5: Minimizing dependence on information made available by \emph{\saf} .}
    %$\mathbb{E}_{\pi_\theta }\!\left[r - \beta %D_{\text{KL}}\left[p_\text{enc}\!\left(Z \mid O, M\right) \mid\mid %p\!\left(Z\mid O\right) \right]\right]\\$
    }
    
\end{algorithm*}
%\end{algorithmic}
%\vspace{-3mm}

%\vspace{-5mm}

\begin{figure}
    \centering
    \includegraphics[width=\linewidth]{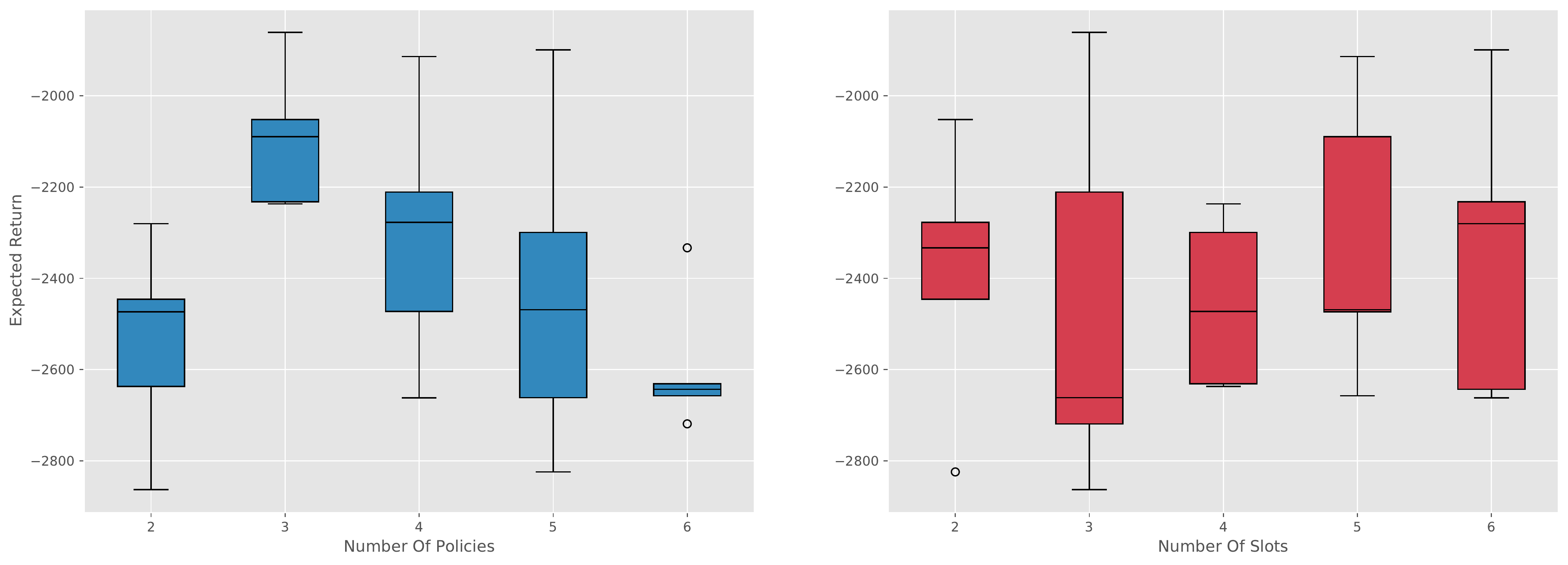}
    
    \caption{Sensitivity of our approach to key hyperparameters: (a.) Dependence on number of policies ($U$) (b.) Dependence on number of \emph{\saf} slots ($l$). We use the Ghost Run environment.}
    \label{fig:hyper}
\end{figure}
%\section{You \emph{can} have an appendix here.}

%You can have as much text here as you want. The main body must be at most $8$ pages long.
%For the final version, one more page can be added.
%If you want, you can use an appendix like this one, even using the one-column format.
%%%%%%%%%%%%%%%%%%%%%%%%%%%%%%%%%%%%%%%%%%%%%%%%%%%%%%%%%%%%%%%%%%%%%%%%%%%%%%%
%%%%%%%%%%%%%%%%%%%%%%%%%%%%%%%%%%%%%%%%%%%%%%%%%%%%%%%%%%%%%%%%%%%%%%%%%%%%%%%

\subsection{MARL environments}

\textbf{GhostRun environment} is a cooperative multi-agent game  adapted from Jiang et al. 2021 (\cite{jiang2021multi}).GhostRun environment consists of multiple agents with partial view of a 2D world of square shape. There are ghosts randomly moving around in the environment. Each agent receive a negative reward of $-10*N_{ghost}$ at each time step where $N_{ghost}$ is the number of ghosts in the agent's partial view of the environment. In addition, there is a step cost of -1 for each agent at each time step. The goal of the game is to maximize the sum of rewards from all agents in the team. We conducted the experiments using fixed number of 100 ghosts and various number of agents.

\textbf{Multi-agent Particle-World Environment (MPE) (results to be added)} was introduced in Lowe et al.(\cite{lowe2017multi}). MPE
consist of various multi-agent games in a 2D world with small particles navigating within a square
box. We consider a cooperative task from the original set called "SimpleSpread". In the task there are various number agents and landmarks. The team reward is calculated by the distance between each landmark and its nearest agents. In this study, 2 landmarks and various number agents are used.

\begin{table}[h]
\caption{Multi-agent reinforcement learning environments used in this work along with the number of agents and the task to solve for each environment.}
\label{env-overview}
\vskip 0.15in
\begin{center}
\begin{small}
\begin{sc}
\begin{tabular}{lll}
\toprule
\textbf{Envs}       & \textbf{N agents} & \textbf{task}     \\
\midrule
\textbf{GhostRun}   & 5                 & Hide from ghosts \\
\textbf{MSTC}       & 3                 & Observe targets   \\
\textbf{PistonBall} & 5                 & Move the ball \\
\bottomrule
\end{tabular}
\end{sc}
\end{small}
\end{center}
\vskip -0.1in
\end{table}

\subsection{Training details}
The optimization algorithm for each agent in our SAF method is PPO. Baselines IPPO, CPPO and MAPPO share the same or similar architectures of actor and critic as in SAF. Hyperparameters such as batch sizes, number of training episodes and learning rates of these PPO backbone are obtained from the original publication \cite{yu2021surprising}.

Architectures and hyperparameters of SAF method were tuned. All the baselines are implemented in a way that their performance either matches or exceeds the results in the original publications if available.

\subsection{Hyperparameters and sensitivity to SAF specific hyperparameters}
Common hyparameters used in SAF and other PPO derived method are shown in table \ref{Hyperparameters}. Other baselines uses hyperparameters from original publications.

In this section we study our approach's senstivity to two key hyperparameters: the number of slots $l=N_{slots}$ and the number of policies $U=N_{policies}$. In our findings that we summarize in Figure \ref{fig:hyper}, we find that \emph{SAF} is robust to variation in the number of slots. For the number of policies however, we find that the performance is best for $N_{policies}=3$ and decreases for bigger values.

\begin{table}[h]
\caption{Common hyperparameters used in SAF and other PPO baselines}
\label{Hyperparameters}
\vskip 0.15in
\begin{center}
\begin{small}
\begin{sc}
\begin{tabular}{ll}
\toprule
\textbf{hyperparameters}       & \textbf{Values}      \\
\midrule
\textbf{Gamma}   & 0.99     \\
\textbf{batch size}       & 15   \\
\textbf{optimizer} & Adam        \\
\textbf{optimizer epsilon} & 0.01        \\
\textbf{weight decay} & 0        \\
\textbf{Feature normalization} & batch norm        \\
\textbf{Reward normalizaiton} & by batch        \\

\bottomrule
\end{tabular}
\end{sc}
\end{small}
\end{center}
\vskip -0.1in
\end{table}

%\end{comment}

\end{document}